\documentclass[10pt,twocolumn,letterpaper]{article}
\usepackage[accsupp]{axessibility}
\usepackage{wacv}
\usepackage{times}
\usepackage{epsfig}
\usepackage{graphicx}
\usepackage{amsmath}
\usepackage{amssymb}

\usepackage{subfigure}
\usepackage{color}
\usepackage{comment}
\usepackage{float}

\usepackage{balance}

\def\wacvPaperID{33} 

\wacvfinalcopy 

\ifwacvfinal
\def\assignedStartPage{1} 
\fi

\ifwacvfinal
\usepackage[breaklinks=true,bookmarks=false]{hyperref}
\else
\usepackage[pagebackref=true,breaklinks=true,colorlinks,bookmarks=false]{hyperref}
\fi

\ifwacvfinal
\else
\fi

\begin{document}


\title{Semantic-Guided Zero-Shot Learning for Low-Light Image/Video Enhancement}

\author{Shen Zheng\\
Wenzhou-Kean University\\
Wenzhou, China\\
{\tt\small zhengsh@kean.edu}

\and
Gaurav Gupta\\
Wenzhou-Kean University\\
Wenzhou, China\\
{\tt\small ggupta@kean.edu}
}

\maketitle
\thispagestyle{empty}

\begin{abstract}
    Low-light images challenge both human perceptions and computer vision algorithms. It is crucial to make algorithms robust to enlighten low-light images for computational photography and computer vision applications such as real-time detection and segmentation. This paper proposes a semantic-guided zero-shot low-light enhancement network (SGZ) which is trained in the absence of paired images, unpaired datasets, and segmentation annotation. Firstly, we design an enhancement factor extraction network using depthwise separable convolution for an efficient estimate of the pixel-wise light deficiency of an low-light image. Secondly, we propose a recurrent image enhancement network to progressively enhance the low-light image with affordable model size. Finally, we introduce an unsupervised semantic segmentation network for preserving the semantic information during intensive enhancement. Extensive experiments on benchmark datasets and a low-light video demonstrate that our model outperforms the previous state-of-the-art. We further discuss the benefits of the proposed method for low-light detection and segmentation. Code is available at https://github.com/ShenZheng2000/Semantic-Guided-Low-Light-Image-Enhancement.
    
\end{abstract}

\begin{figure}[t]
\centering
\subfigure[Dark]{
\includegraphics[width=3.7cm]{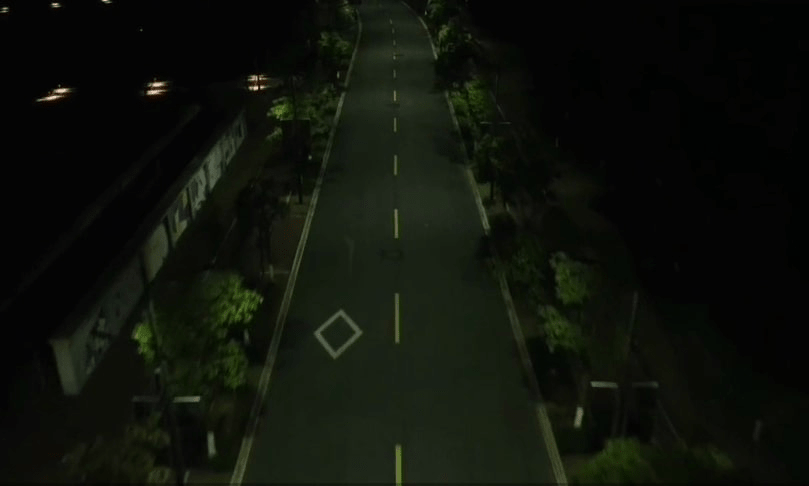}
}
\subfigure[Retinex \cite{wei2018deep}]{
\includegraphics[width=3.7cm]{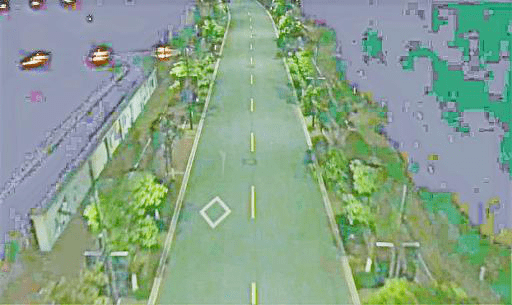}
}
\subfigure[KinD \cite{zhang2019kindling}]{
\includegraphics[width=3.7cm]{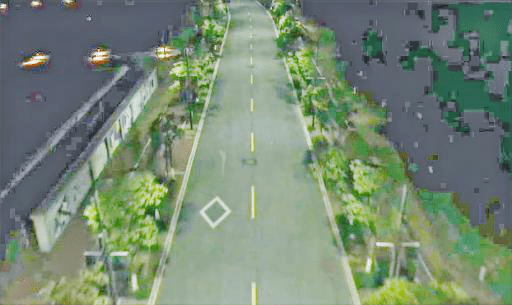}
}
\subfigure[EnlightenGAN \cite{jiang2021enlightengan}] {
\includegraphics[width=3.7cm]{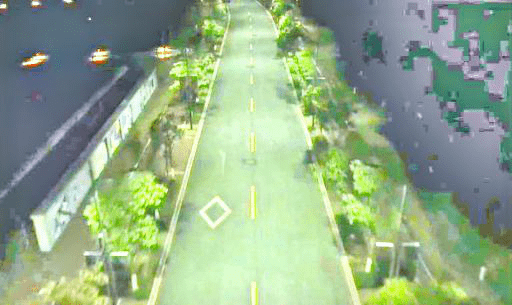}
}
\subfigure[Zero-DCE \cite{guo2020zero}]{
\includegraphics[width=3.7cm]{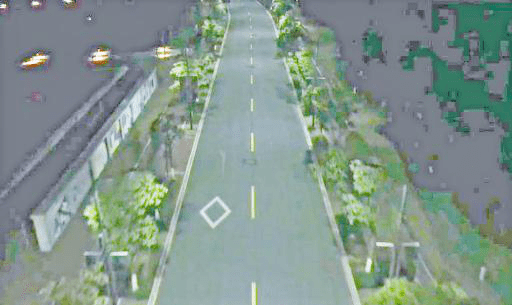}
}
\subfigure[Ours]{
\includegraphics[width=3.7cm]{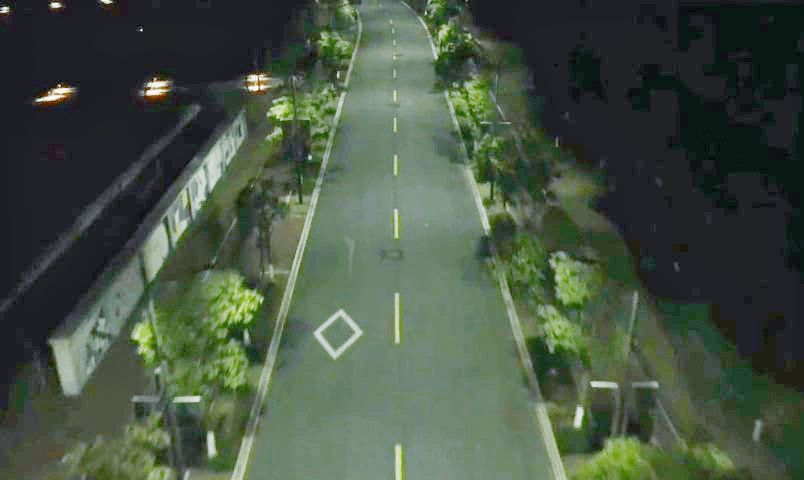}
}
\caption{The enhancement result on a nighttime aerial video frame. Our proposed model has excellent perceptual quality in terms of exposure, contrast, color, and edge information. In comparison, other models either fail to enhance the dark regions or generate unpleasant noise, blur or artifacts.}
\label{sample}
\end{figure}

\begin{figure*}
\centering
\includegraphics[width=16cm]{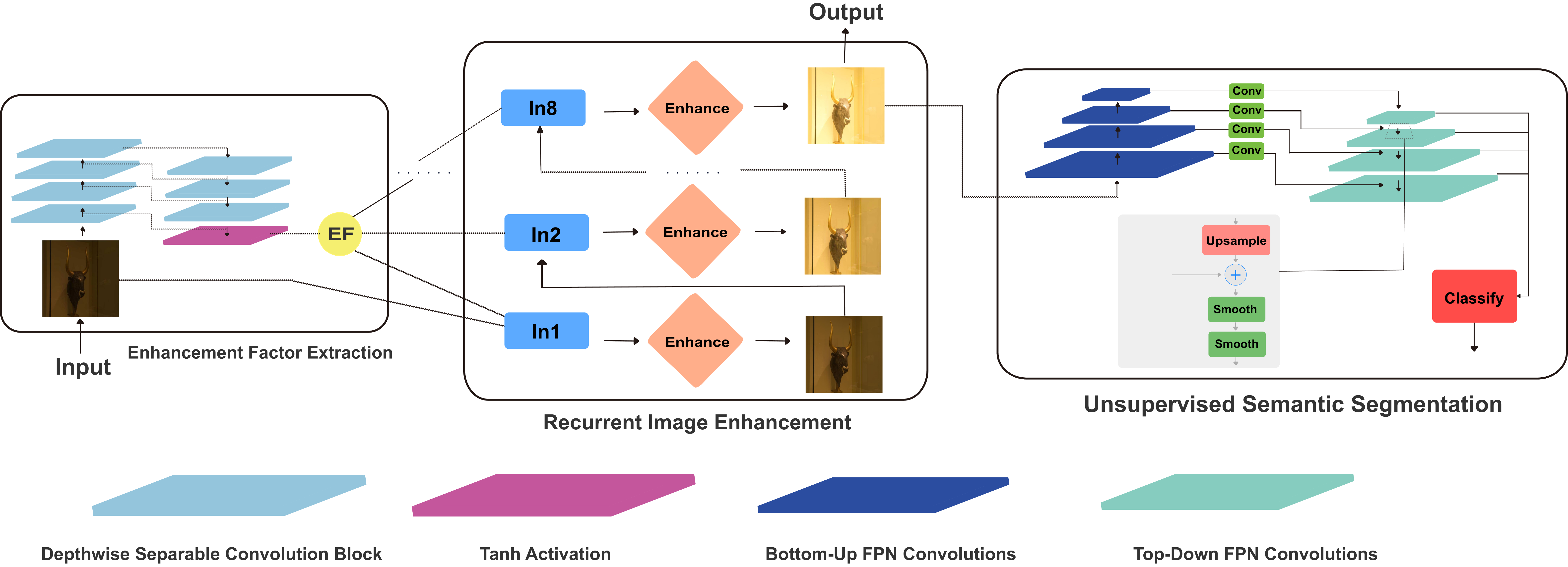}
\caption{The proposed model architecture. Our model consists of a three stage network: EFE for estimating the light enhancement factor, RIE for progressively lighten the image, and USS for segment the enhanced image. During training, both RIE and USS have frozen parameters and they output the loss to update EFE. During testing, EFE and RIE are used sequentially to enhance an low-light image.}
\label{ModelArchs}
\end{figure*}


\section{Introduction}
\par
\footnote{This work is supported by the research funding from
Wenzhou-Kean University with project number SpF2021011. We thank Changjie Lu for his help on model architecture design.} Low-light images degraded due to environmental or technical restraints suffer from various problems such as under-exposure and high ISO noise. As a result, those images are prone to have degraded features and contrast, which harm the low-level perceptual quality and deteriorate high-level computer vision tasks relying on accurate semantic information. It is necessary to improve the visual quality and to enhance the generalizability of the advanced vision algorithms.

One plausible way to increase brightness at low-light conditions is to use higher ISO or more extended exposure time. Nevertheless, those strategies respectively intensify noises and introduce motion blur \cite{chen2018learning}. The other reasonable approach is to use modern software like Photoshop or Lightroom for light adjustment. However, these software requires artistic skills and are inefficient for large-scale datasets with diverse illumination conditions.
  
Traditional low-light image enhancement methods mostly involves Histogram Equalization \cite{ibrahim2007brightness, wang2007fast} and Retinex theory \cite{land1977retinex, wang2013naturalness, fu2015probabilistic, Fu_2016_CVPR, guo2016lime}. Although these methods can generate encouraging perceptual qualities in some situations, their performances depend on manually selected priors and hand-crafted regularization which are difficult to tune. Furthermore, the long inference time resulting from the intricate optimization process makes them unfitting for real-time tasks.

Deep learning based low-light image enhancement methods have recently received much attention due to their compelling efficiency, accuracy, and robustness \cite{li2021lighting}. Supervised methods \cite{lore2017llnet, wei2018deep, zhang2019kindling, wang2019underexposed} have the highest scores at some benchmark datasets \cite{wang2013naturalness, guo2016lime, ma2015perceptual, lee2012contrast} with their excellent image-to-image mapping abilities. However, they require paired training images (i.e., low/normal light pairs), which either need expensive retouch or demand unfeasible image capture with the same scene but different lighting conditions. On the other hand, Unsupervised methods \cite{jiang2021enlightengan} require only an unpaired dataset for training. Nonetheless, the data bias from the manually selected datasets restricts their generalization ability. Zero-shot learning \cite{guo2020zero, li2021learning} methods eliminate the need for both paired images and unpaired dataset. However, they ignore the semantic information, which is shown by \cite{mottaghi2014role, gonzalez2018objects, liu2019visual} to be crucial for high-level vision tasks. As a result, their enhanced images are in sub-optimal visual quality. Fig. \ref{sample} reveals the limitations of the previous researches.

To address the limitations discussed above, we present a semantic-guided zero-shot framework for low-light image enhancement (Fig. \ref{ModelArchs}). As we focus on low-light image/video enhancement, we first design a light-weight enhancement factor extraction (EFE) network with depthwise separable convolution \cite{howard2017mobilenets} and symmetric skip connections. The EFE is highly adaptive and can leverage the spatial information of the low-light images to monitor the subsequent image enhancement. To perform image enhancement with affordable model size, we then introduce a recurrent image enhancement (RIE) network which utilizes both the low-light image and the enhancement factor from EFE as its input. The RIE is able to progressively enhance the images, using the previous stage's output as the input for the subsequent recurrent stage. Aiming to preserve the semantic information during the enhancement process, we finally propose an unsupervised semantic segmentation (USS) network requiring no expensive segmentation annotation. The USS receives the enhanced image from RIE and utilizes feature pyramid network \cite{lin2017feature} to calculate the segmentation loss. The segmentation loss merges with other non-reference loss functions as the total loss, which updates the parameters of EFE during training.

The contributions of the proposed work are summarized as follows: 
\begin{itemize}
    \item We propose a new semantic-guided zero-shot low-light image enhancement network. To the best of our knowledge, we are the first to fuse high-level semantic information into low-level image enhancement with the absence of paired images, unpaired datasets, or segmentation labels.
    \item We develop a light-weight convolutional neural network to automatically extract the enhancement factor which record the pixel-wise light deficiency of an low-light image.
    \item We design an recurrent image enhancement strategy with five non-reference loss functions to boost our model's generalization ability to images of diverse lighting conditions. 
    \item We conduct extensive experiments to demonstrate the superiority of our model in both qualitative and quantitative metrics. \textbf{Our model is ideal for low-light video enhancement because it can process 1000 images of size $\textbf{1200}\times\textbf{900}$ within 1 second on a single GPU}.
\end{itemize}


\begin{figure}[t]
\centering
\subfigure{
\includegraphics[width=3.5cm]{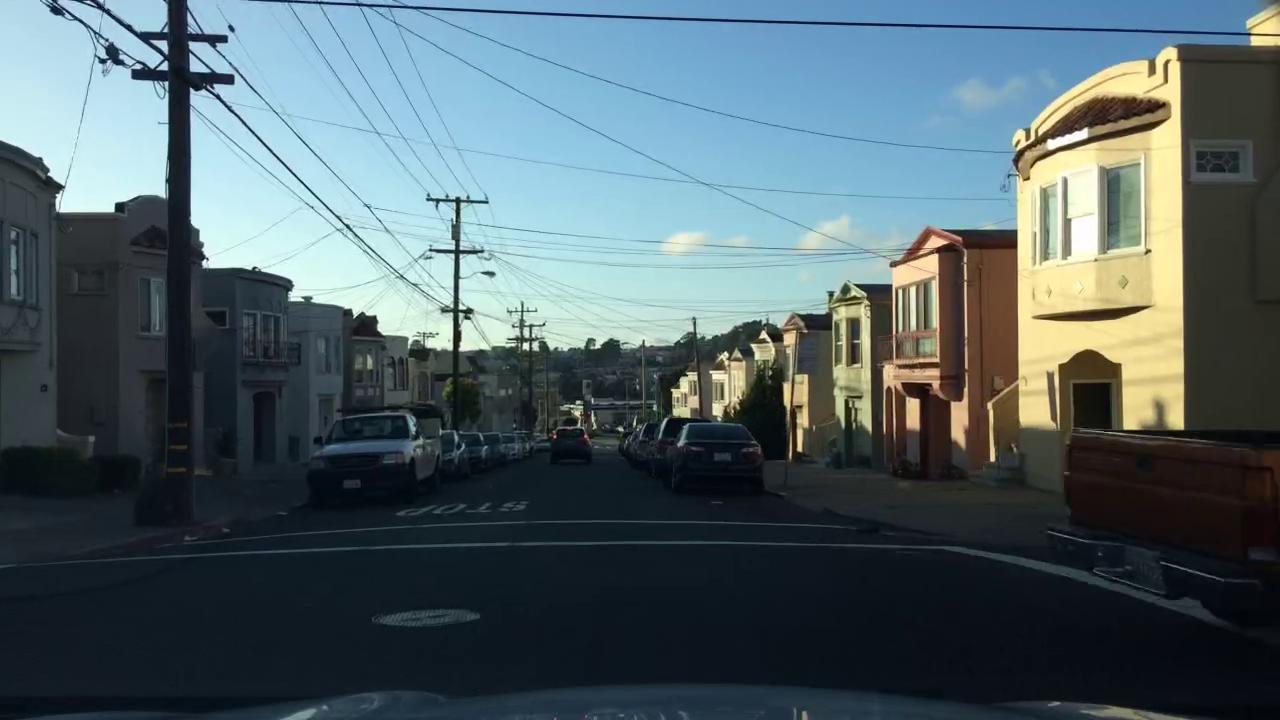}
}
\subfigure{
\includegraphics[width=3.5cm]{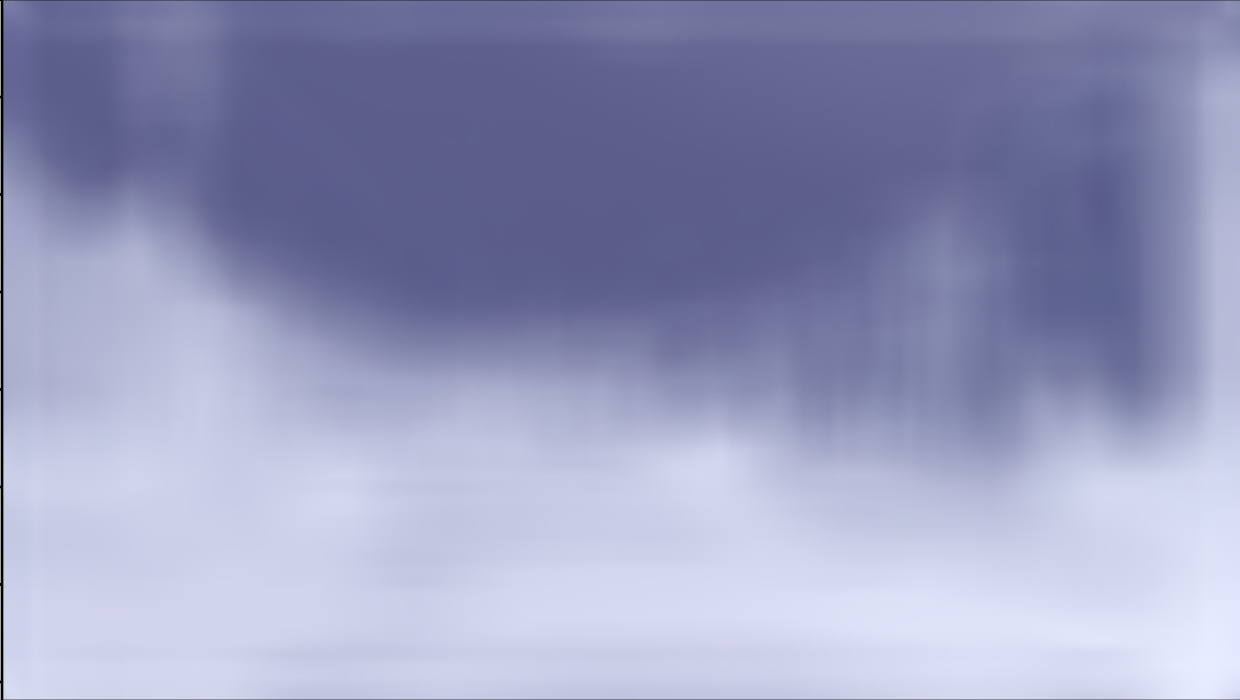}
}
\subfigure{
\includegraphics[width=3.5cm]{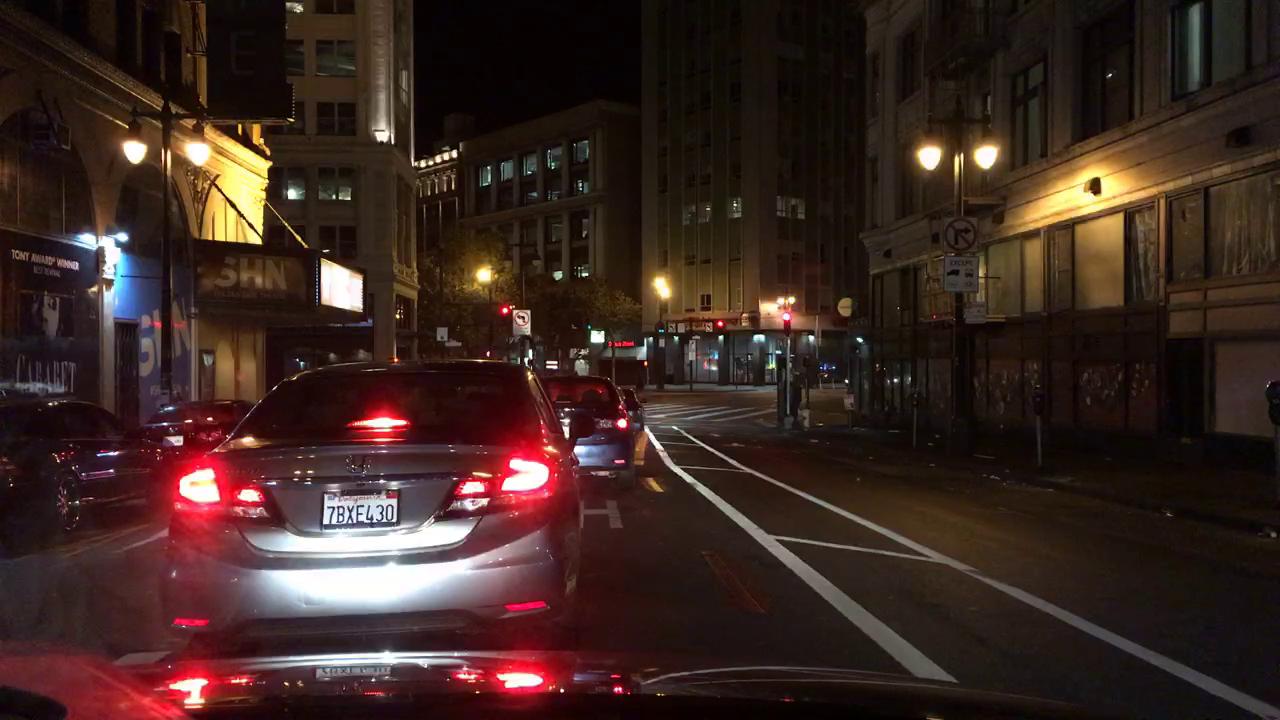}
}
\subfigure{
\includegraphics[width=3.5cm]{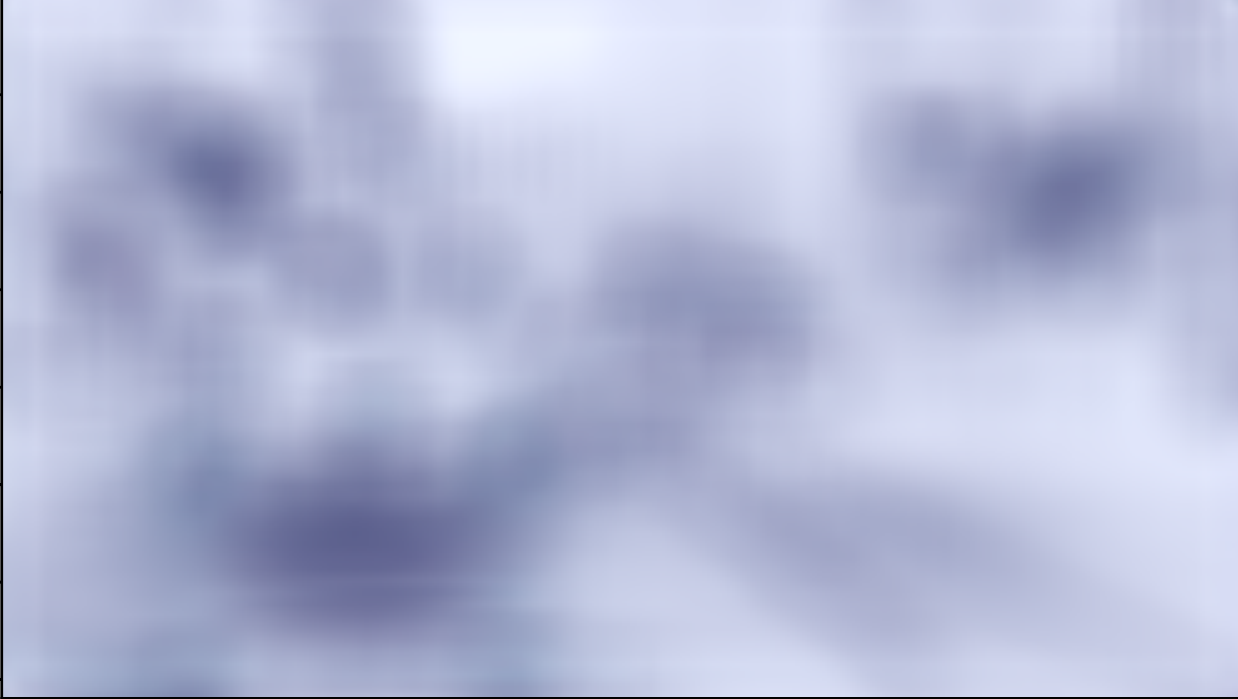}
}
\caption{Enhancement factor visualization. Left column: Low-Light Images. Right column: Corresponding Enhancement Factor. Darker region indicates lower values for the enhancement factor.}
\label{EnhanceFactor}
\end{figure}

\section{Related Work}

\noindent
\textbf{Traditional Low-Light Image Enhancement}
Traditional low-light Image Enhancement mainly consists of histogram equalization (HE-based) methods and Retinex-based methods. HE-based image enhancement methods have been widely applied in the early years. BPDHE \cite{ibrahim2007brightness} proposes a brightness preserving dynamic histogram equalization method that can maintain the mean intensity of the low-light image in its enhanced version. WTHE \cite{wang2007fast} introduces a contrast enhancement method that performs weighting and thresholding on the histogram of an image before the histogram equalization operation.  


Recently, many Retinex-based methods have been designed for low-light image enhancement. NPE \cite{wang2013naturalness} proposes a non-uniform, naturalness-preserving enhancement method to balance image details and naturalness. PIE \cite{fu2015probabilistic} presents a probabilistic enhance approach which exploits concurrent estimation of illumination and reflectance. LIME \cite{guo2016lime} estimate a coarse illumination map finding the maximum value in the R, G, B channel and then improve that coarse map using a structure prior. 


Unlike conventional methods, our model uses a light-weight convolutional neural network to automatically extract the enhancement factor that learns the enlightenment requirement from the low-light images. That design allows the recurrent image enhancement network to run in linear complexity yet still achieving compelling results.

\noindent
\textbf{Deep Low-Light Image Enhancement} 
Deep learning based low-light image enhancement methods can be mainly classified into supervised learning, unsupervised learning, and zero-shot learning. The pioneering supervised low-light enhancement method LLNet \cite{lore2017llnet} presents a noise-robust autoencoder-based way to enlighten images with minimum pixel-level saturation. Retinex \cite{wei2018deep} considers Retinex theory, integrating a decomposition network and an illumination adjustment network that learns from paired low/normal light pictures. The similar work KinD \cite{zhang2019kindling} additionally introduce degradation removal in the reflectance.


Unsupervised methods avoid the tedious work for preparing paired training images. EnlightenGAN \cite{jiang2021enlightengan} is the first low-light image enhancement method trained without paired data. It utilizes an attention-based multi-scale discriminator with self-regularized loss functions. Zero-shot learning eliminates the need for both paired images and unpaired datasets. Zero-DCE \cite{guo2020zero,li2021learning} designs a lightweight network for light-enhancement curves approximation and use non-reference loss functions to enhance the low-light images.

Unlike other deep low-light image enhancement methods, our model exploits the high-level semantic information with a pretrained segmentation network requiring no segmentation label. That design allows us to preserve an essential amount of semantic information without significantly increasing the computational complexity.

\begin{figure}[t]
\centering
\subfigure[$\left|x_{r}\right|$ = 0.2]{
\includegraphics[width=2.5cm]{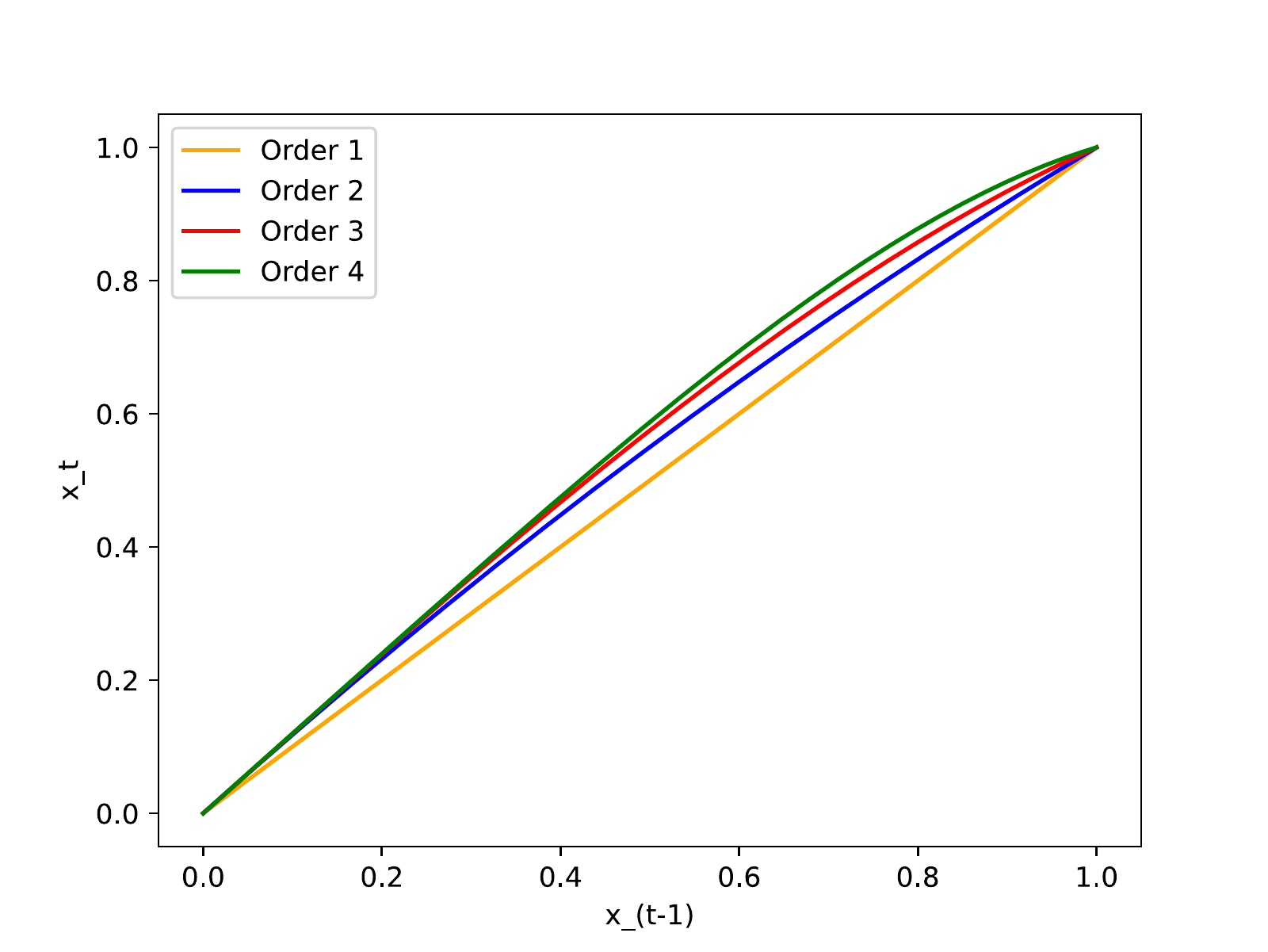}
}
\subfigure[$\left|x_{r}\right|$ = 0.5]{
\includegraphics[width=2.5cm]{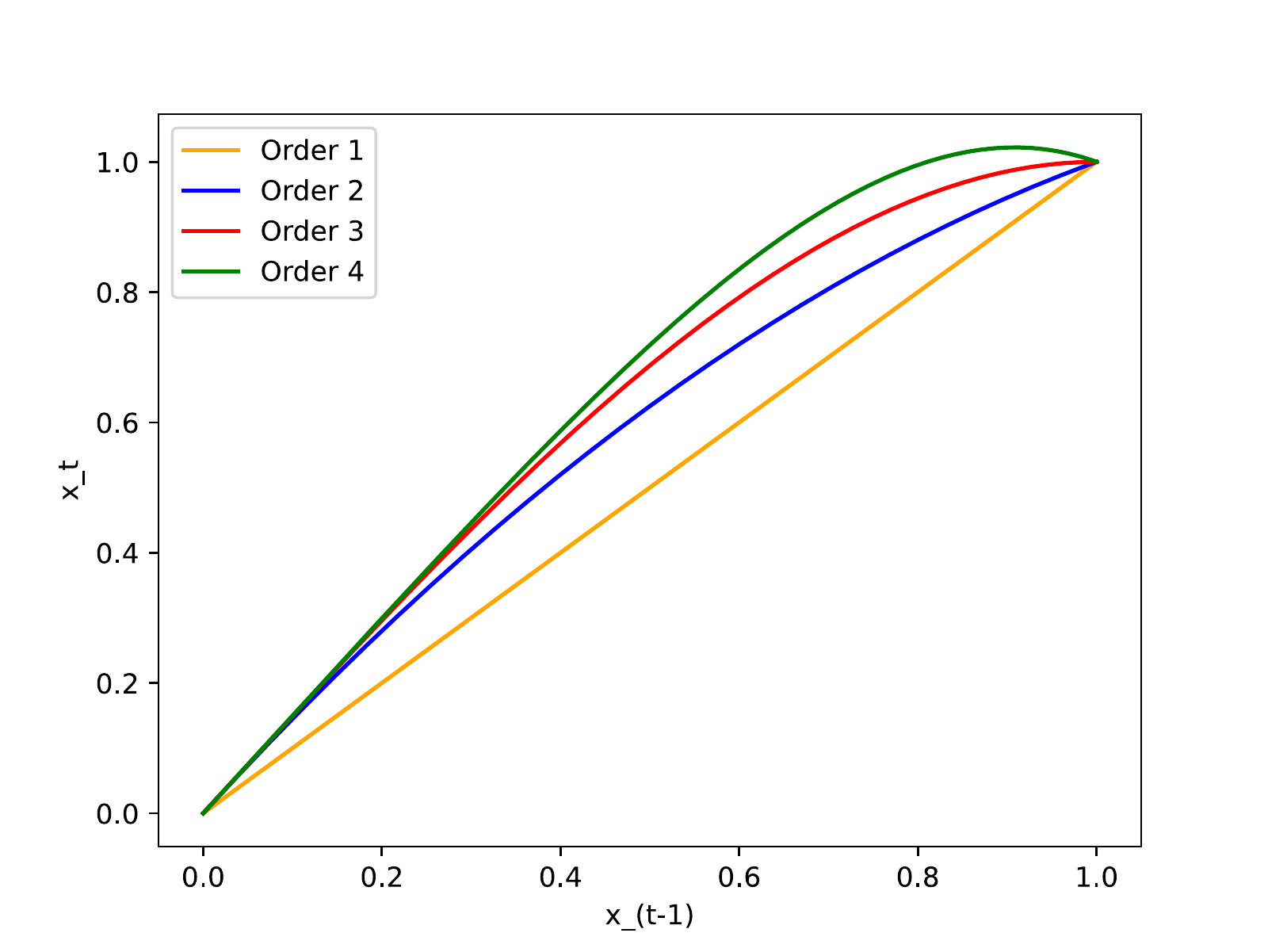}
}
\subfigure[$\left|x_{r}\right|$ = 0.8]{
\includegraphics[width=2.5cm]{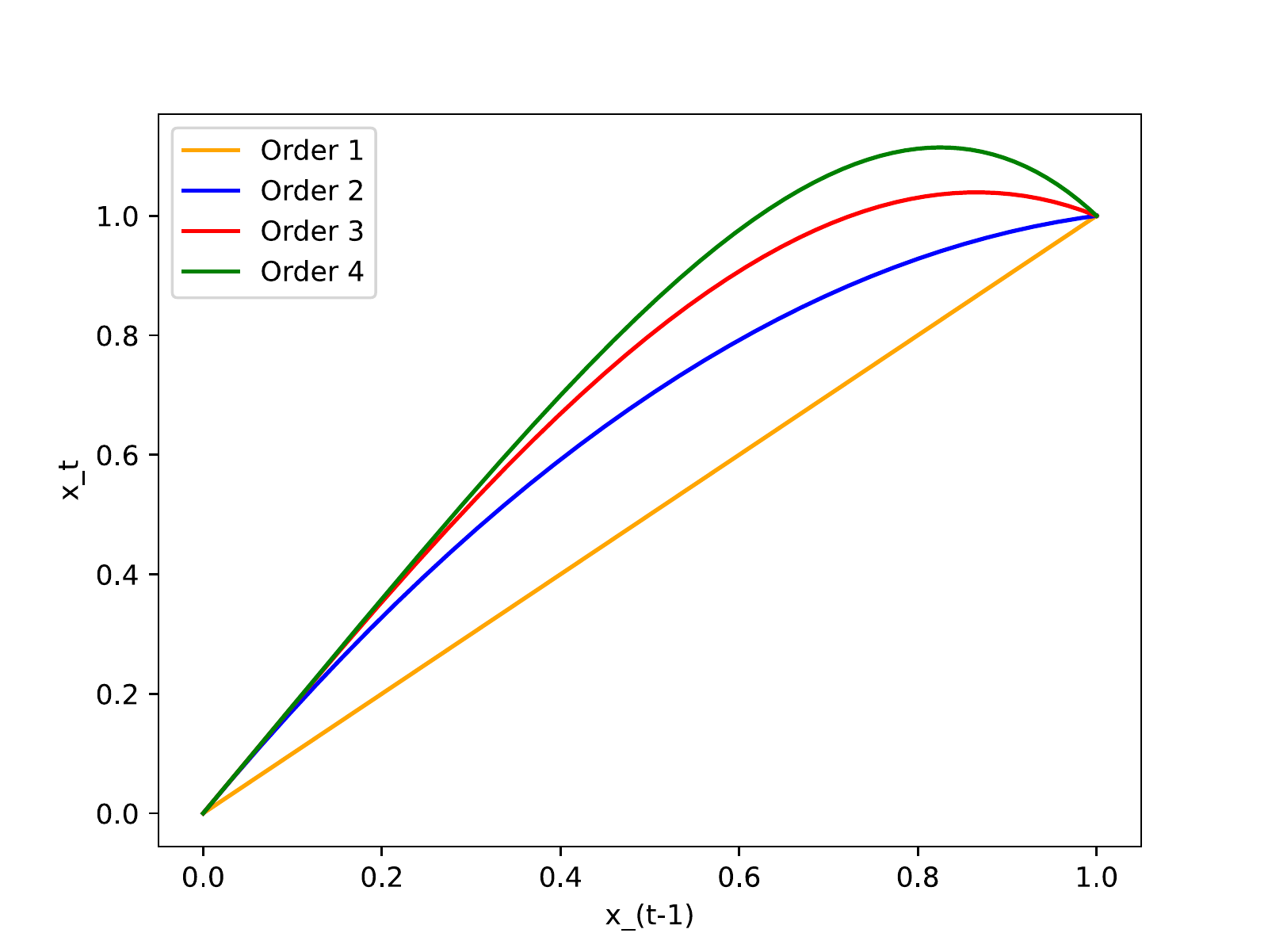}
}
\caption{Recurrent image enhancement illustration with different enhancement factor $x_{r}$ and different Order. The horizontal axis $x_{t-1}$ refers to a pixel's value in the $(t-1)$ stage, whereas the vertical axis $x_{t}$ refers to that pixel's value in the $t$ stage. All pixel value is scaled to [0, 1]. Greater $\left|x_{r}\right|$ indicates a more intense enhancement.}
\label{AbaCurve}
\end{figure}

\section{Proposed Method}


\subsection{Enhancement Factor Extraction Network}
The enhancement factor extraction (EFE) aims to learns the pixel-wise light deficiency of a low-light image and records that information in a enhancement factor. Inspired by the architecture of U-Net \cite{ronneberger2015u}, EFE is a fully convolutonal neural network with symmetric skips connections, which means that it can address input images of arbitrary size. No batch normalization or up/downsampling is adopted since they will damage the spatial coherence of the enhanced image \cite{ulyanov2016instance, johnson2016perceptual, isola2017image}. Each convolution block in EFE consists of a $3\times3$ depthwise separable convolution layer and a subsequent ReLU \cite{nair2010rectified} activation layer. The last convolution block reduces the channel numbers from 32 to 3 and output the enhancement factor $x_r$ via the Tanh activation. Fig. \ref{EnhanceFactor} visualizes the enhancement factor extracted from 2 low-light images. It is evident that brighter regions in the low-light image corresponds to lower values in the enhancement factor, and vice verse.

\begin{figure}[t]
\centering
\includegraphics[width=7cm]{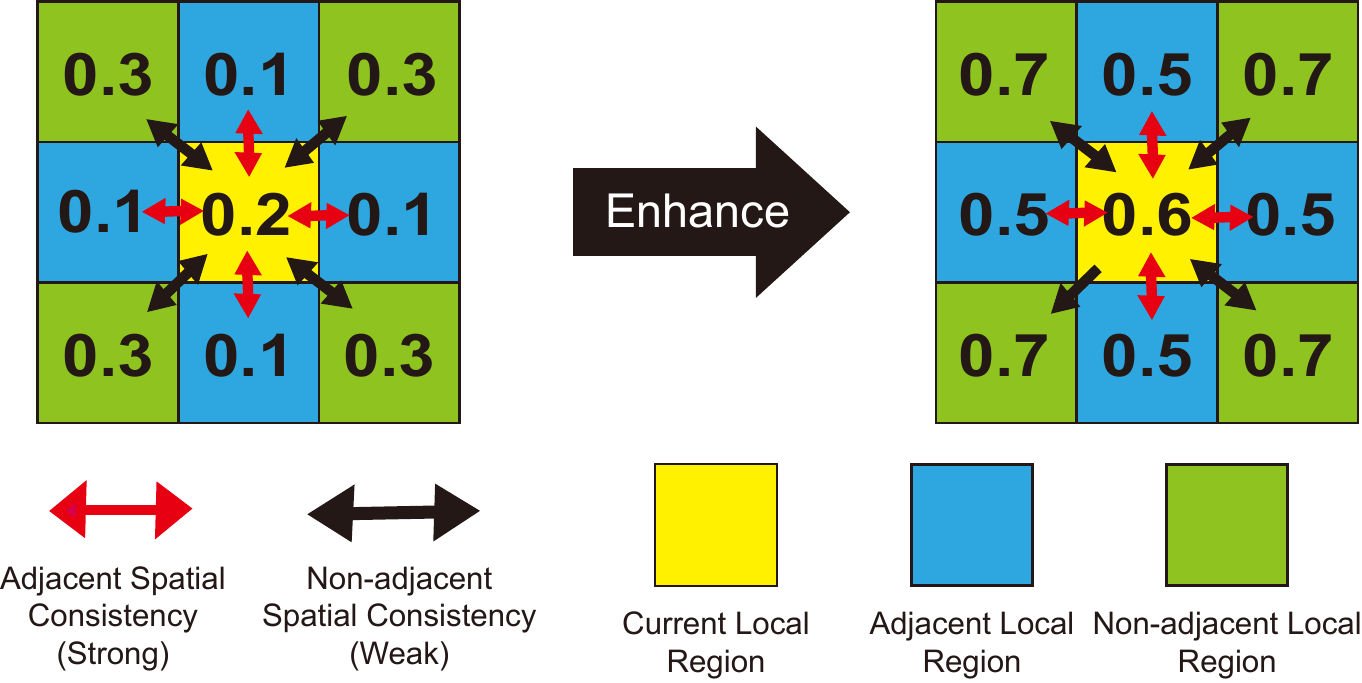}
\caption{Illustration of spatial consistency loss $L_{s p a}$, assuming the pixel value of the input image is scaled to [0, 1]. $L_{s p a}$ encourages the connection between neighboring local regions when the enhancement is increasing all pixel values.}
\label{SpaLossFig}
\end{figure}

\subsection{Recurrent Image Enhancement Network}
Inspired by the success of recurrence \cite{ren2019low, zhu2020eemefn, li2021low} and light enhancement curve \cite{yuan2012automatic, guo2020zero} in low-light image enhancement, we build a recurrent image enhancement (RIE) network to enhance the low-light image according to the enhancement factor and then output the enhanced image. Each recurrence considers the previous stage's output and the enhancement factor as its input. The recurrent enhancement process is:

\begin{equation}
x_t = x_{t-1} + x_r * (x_{t-1}^\text{Order} - x_{t-1})
\end{equation}

\noindent
where $x$ is the output, $x_r$ is the enhancement factor and $t$ is the recurrence step. The next step is to decide the optimal Order to enlight the image. Since the recurrent network should be simple for differentiation and should be effective for progressive lightening, we only consider positive integers for Order. With this in mind, we plot the recurrent image enhancement with respect to different $x_r$ and Order in Fig. \ref{AbaCurve}. When Order is 1, the pixel value is insensitive to $x_r$ and is the same as the previous stage. When Order equals 3 or 4, the pixel value approaches or even exceed 1.0, making a image looks too bright. In comparison, Order of 2 grants the most robust enhancement in recurrence.

\subsection{Unsupervised Semantic Segmentation Network}
The Unsupervised Semantic Segmentation (USS) Network aims at accurate pixel-wise segmentation of the enhanced image which preserve the semantic information during progressive image enhancement. Similar to \cite{fan2018segmentation, liu2017image, wang2019segmentation, guo2020high}, we freeze all the layers for segmentation network during training. Here, we use two pathways, including the bottom-up pathway which uses ResNet-50 \cite{he2016deep} with ImageNet \cite{deng2009imagenet} weights, and the top-down pathway which uses Gaussian initialization with a mean of 0 and a standard deviation of 0.01. Both pathways have four convolution blocks which is connected to each other through lateral connections. The choice of weight initialization will be explained in the ablation study.

The enhanced image from RIE will first enter the bottom-up pathway for feature extraction. The top-down pathway then transforms the high-semantic layers into high-resolution ones for spatial-aware semantic segmentation. Each convolution block in the top-down approach performs bi-linear upsampling on the image and concatenates it with the lateral outcome. Two smooth layers with $3\times3$ convolution are applied after the concatenation for better perceptual quality. Finally, we concatenate the result of each block in the top-down pathway and calculate the segmentation.

\subsection{Loss Functions}
We adopt five non-reference loss functions, including $L_{s p a}$, $L_{rgb}$, $L_{bri}$, $L_{t v}$ ,and $L_{sem}$. We do not consider content loss or perceptual loss \cite{ma2015perceptual} due to the unavailability of paired training images.

\noindent
\textbf{Spatial Consistency Loss} This Spatial Consistency loss helps to maintain the spatial consistency between the low-light image and the enhanced image by conserving the neighbor pixels' differences during enhancement. Unlike \cite{guo2020zero, li2021learning} that only consider adjacent cells, we also include the spatial coherence with non-adjacent neighbors (See Fig. \ref{SpaLossFig}). The spatial consistency loss is:

\begin{equation}
\begin{split}
L_{s p a}=\frac{1}{A} \sum_{i=1}^{A} [
\sum_{j \in \phi(i)}\left(\left|\left(Y_{i}-Y_{j}\right)\right|-\left|\left(I_{i}-I_{j}\right)\right|\right)^{2}
+ \\ \alpha * 
\sum_{k \in \psi(i)}\left(\left|\left(Y_{i}-Y_{k}\right)\right|-\left|\left(I_{i}-I_{k}\right)\right|\right)^{2} ]
\end{split}
\end{equation}

\noindent
where $Y$ and $I$ are the mean pixel value in a $A \times A$ local region in an enhanced image and the low-light image, respectively. $A$ is the side of the local regions which we set to 4 according to the ablation study. $\phi(i)$ is the four adjacent neighbors (top, down, left, right), and $\psi(i)$ is the four non-adjacent neighbors (top left, top right, lower left, and lower right). $\alpha$ is 0.5 since the weight of the non-adjacent neighbors is less important.

\noindent
\textbf{RGB Loss} The color loss \cite{wang2019underexposed, zhang2019kindling, guo2020zero} reduces color incorrectness in the enhanced picture by bridging different color channels. We adopt Charbonnier loss which helps high-quality image reconstruction \cite{lai2018fast, jiang2020multi}. The RGB loss is:

\begin{equation}
\begin{split}
L_{rgb}=\sum_{\forall(i, j) \in \zeta}\sqrt{\left(\left(Y^i\right)-
\left(Y^j\right)\right)^{2}+\varepsilon^{2}}, \\
\zeta = \{(R, G),(R, B),(G, B)\}
\end{split}
\end{equation}

\noindent
where $\varepsilon$ is a penalty term that is empirically set to $10^{-6}$ for training stability.

\begin{figure}[t]
\centering
\subfigure{
\includegraphics[width=1.4cm]{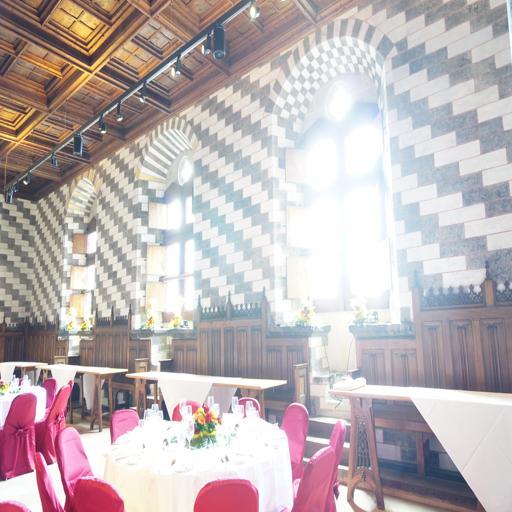}
}
\subfigure{
\includegraphics[width=1.4cm]{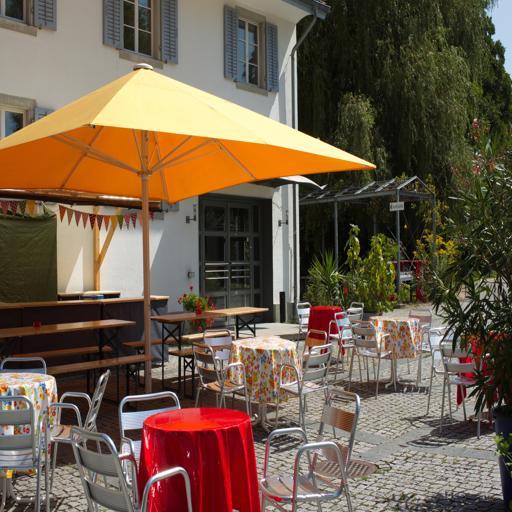}
}
\subfigure{
\includegraphics[width=1.4cm]{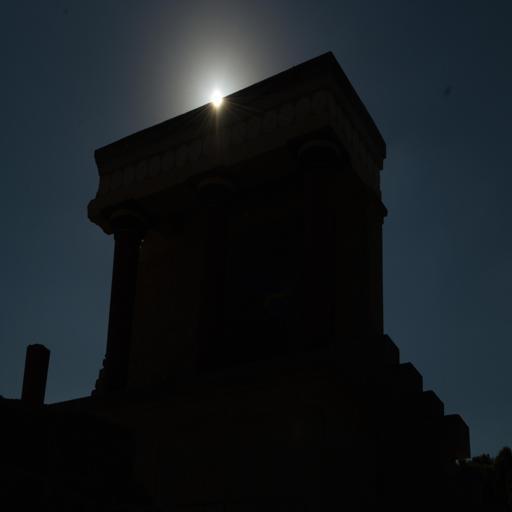}
}
\subfigure{
\includegraphics[width=1.4cm]{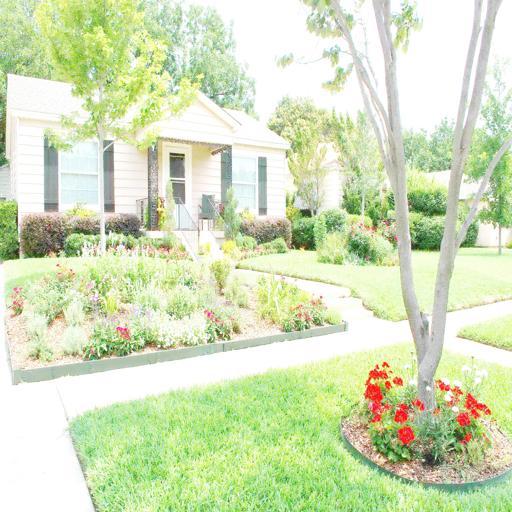}
}
\subfigure{
\includegraphics[width=1.4cm]{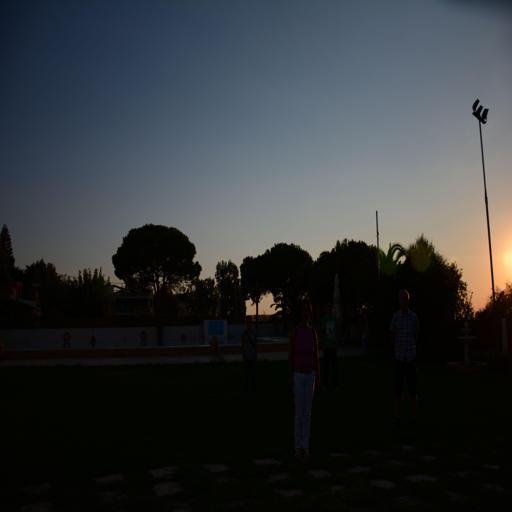}
}
\caption{Sample training images. Our training dataset consists of images of different backgrounds and diverse illumination conditions.}
\label{TrainData}
\end{figure}

\noindent
\textbf{Brightness Loss}
Inspired by \cite{lv2018mbllen, wang2019underexposed, guo2020zero}, we design a brightness loss to constrains the under/over-exposure in an image. The loss measures the L1 difference between the average pixel value of a specific region to a predefined exposure level $E$. The brightness loss is:

\begin{equation}
L_{bri}=\frac{1}{A} \sum_{a=1}^{A}\left|Y_{a}-E\right|
\end{equation}

\noindent
where $E$ is the ideal image exposure level which is set to 0.60 according to the ablation study.

\noindent
\textbf{Total Variation Loss}
The total variation loss \cite{chen2010adaptive} measures the difference between the neighboring pixels in an image. We use total variation loss here to reduce noise and to increase image smoothness. Unlike prior low-light image enhancement works \cite{wei2018deep, zhang2019kindling, wang2019underexposed, guo2020zero}, we additionally consider inter-channel (R, G, and B) relations in the loss to improve the color brightness. Our total variation loss is:

\begin{equation}
\begin{split}
L_{t v}=\frac{1}{CHW} \sum_{c=1}^{C} \sum_{h=1}^{H} \sum_{w=1}^{W}\left[(\nabla_{x} Y_{c,h,w})^2+
(\nabla_{y} Y_{c,h,w})^2 \right]
\end{split}
\end{equation}

\noindent
where $C$, $H$ and $W$ represents channel, height and width of a image, respectively. $\nabla_{x}$ and $\nabla_{y}$ is the horizontal and vertical gradient operations, respectively.

\noindent
\textbf{Semantic Loss}
The semantic loss helps to maintain the semantic information of an image during enhancement. We refer to the focal loss \cite{lin2017focal} for writing our cost function. Recommended by the ablation study, our semantic loss require no segmentation label and only a pre-initialized model. The semantic loss is:

\begin{equation}
L_{sem}=\frac{1}{H W} \sum_{1 \leq i \leq H, 1 \leq j \leq W}-\beta\left(1-p_{i, j}\right)^{\gamma} \log p_{i, j}
\end{equation}

\noindent
where $p$ is the segmentation network's estimated class probability for a pixel. Inspired by \cite{fan2018segmentation}, we chose the focal coefficient $\beta$ and $\gamma$ as 1 and 2, respectively. 

\noindent
\textbf{Total Loss}
The total loss function can be summarized as:
\begin{equation}
\begin{split}
L_{total}=\lambda_{s p a} * L_{s p a}+
\lambda_{rgb} * L_{rgb}+
\lambda_{bri} * L_{bri}+ \\
\lambda_{t v} * L_{t v}+ 
\lambda_{sem} * L_{sem}
\end{split}
\end{equation}

\noindent
Here, we set $\lambda_{s p a}= \lambda_{rgb}=\lambda_{bri}=\lambda_{t v}=1$ and $\lambda_{sem}=0.1$.

\section{Experiments}

\subsection{Implementation Details}
We select 2002 images of different exposure levels and resize them to $512\times512$ (See Fig. \ref{TrainData}) for our model training. The proposed model is trained with Pytorch \cite{paszke2019pytorch} on a single NVIDIA 2080 Ti GPU for 100 epochs using the Adam \cite{kingma2014adam} optimizer with an initial learning rate of 0.0001. The batch size is 6, which takes around 3 hours to converge. Besides, we clip gradient norm to be within 0.1. For initialization of the EFE, we use a normally distributed weight with zero mean and a standard deviation of 0.02.

\subsection{Evaluation Dataset}

We consider two traditional methods PIE \cite{fu2015probabilistic} and LIME \cite{guo2016lime}, three supervised deep learning methods Retinex \cite{wei2018deep}, MBLLEN \cite{lv2018mbllen} and KinD \cite{zhang2019kindling}, one unsupervised method EnlightenGAN \cite{jiang2021enlightengan}, and one zero-shot learning method Zero-DCE\cite{guo2020zero} for model comparison.


Our datasets for comparison includes NPE \cite{wang2013naturalness}, LIME \cite{guo2016lime}, MEF \cite{ma2015perceptual}, DICM \cite{lee2012contrast}, VV\footnote{https://sites.google.com/site/vonikakis/datasets} and LOL \cite{wei2018deep}. Since this paper aims at enhancement of low-light RGB images, we do not include raw datasets such as MIT-Adobe FiveK \cite{bychkovsky2011learning} or SID \cite{chen2018learning}. Instead, moving towards task-driven low-light image enhancement, we additionally select 100 low-light images from BDD10K \cite{yu2018bdd100k} and name it DarkBDD. Besides, we use gamma correction on 150 images from the CityScape \cite{cordts2016cityscapes} dataset to synthesize a new dataset called DarkCityScape. 

For evaluating the model performance, we use reference metrics including Peak Signal-to-Noise Ratio (PSNR), Structure Similarity Index (SSIM) and Mean Square Error (MSE), and non-reference metric including Unified No-reference Image Quality and Uncertainty Evaluator (UNIQUE) \cite{zhang2021uncertainty} and Blind/Referenceless Image Spatial Quality Evaluator (BRISQUE) \cite{mittal2012no}. The overall description is in Table \ref{dataset}.

\begin{table}[t]
\small
\centering
\begin{tabular}{l|l l l l}
\hline
\textbf{Name} & \textbf{Number} & \textbf{Format} & \textbf{Type} & \textbf{Metric }    \\ \hline
NPE\cite{wang2013naturalness}           & 10              & RGB    & Real & U, B  \\ 
LIME\cite{guo2016lime}          & 84& RGB    & Real & U, B  \\ 
MEF\cite{ma2015perceptual}           & 17              & RGB    & Real & U, B  \\
DICM\cite{lee2012contrast}          & 64              & RGB    & Real & U, B  \\ 
VV            & 24              & RGB    & Real & U, B  \\ 
LOL\cite{wei2018deep}           & 15              & RGB    & Real & P, S, M \\
DarkBDD       & 100             & RGB    & Real & U, B  \\ 
DarkCityScape & 150             & RGB    & Synthetic  & P, S, M \\ \hline
\end{tabular}
\caption{Dataset description. Where U, B stands for UNIQUE and BRISQUE, and P, S, M stands for PSNR, SSIM, MSE, respectively. }
\label{dataset}
\end{table}

\begin{figure*}
\centering

\subfigure{
\includegraphics[width=1.9cm]{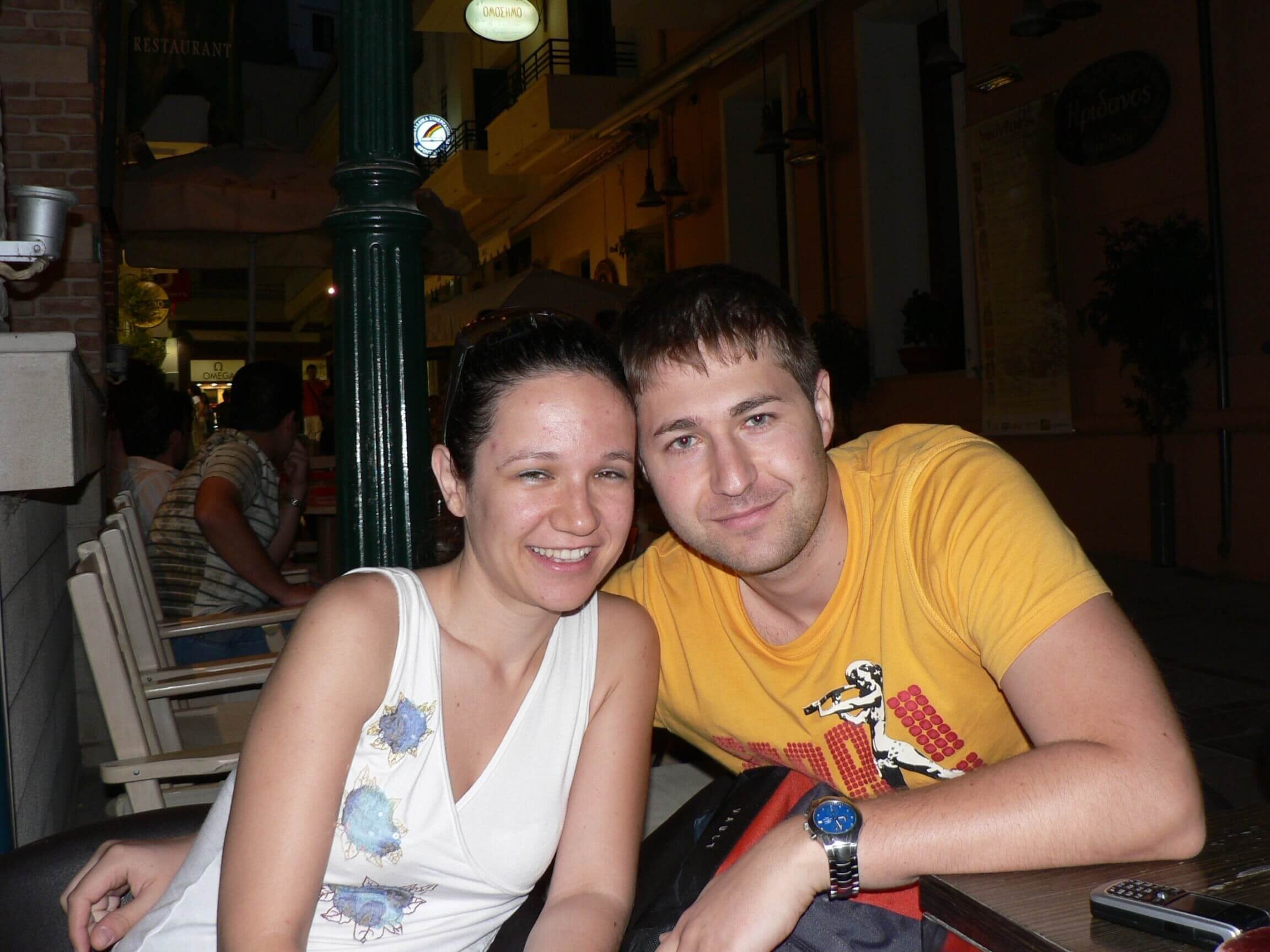}
}
\subfigure{
\includegraphics[width=1.9cm]{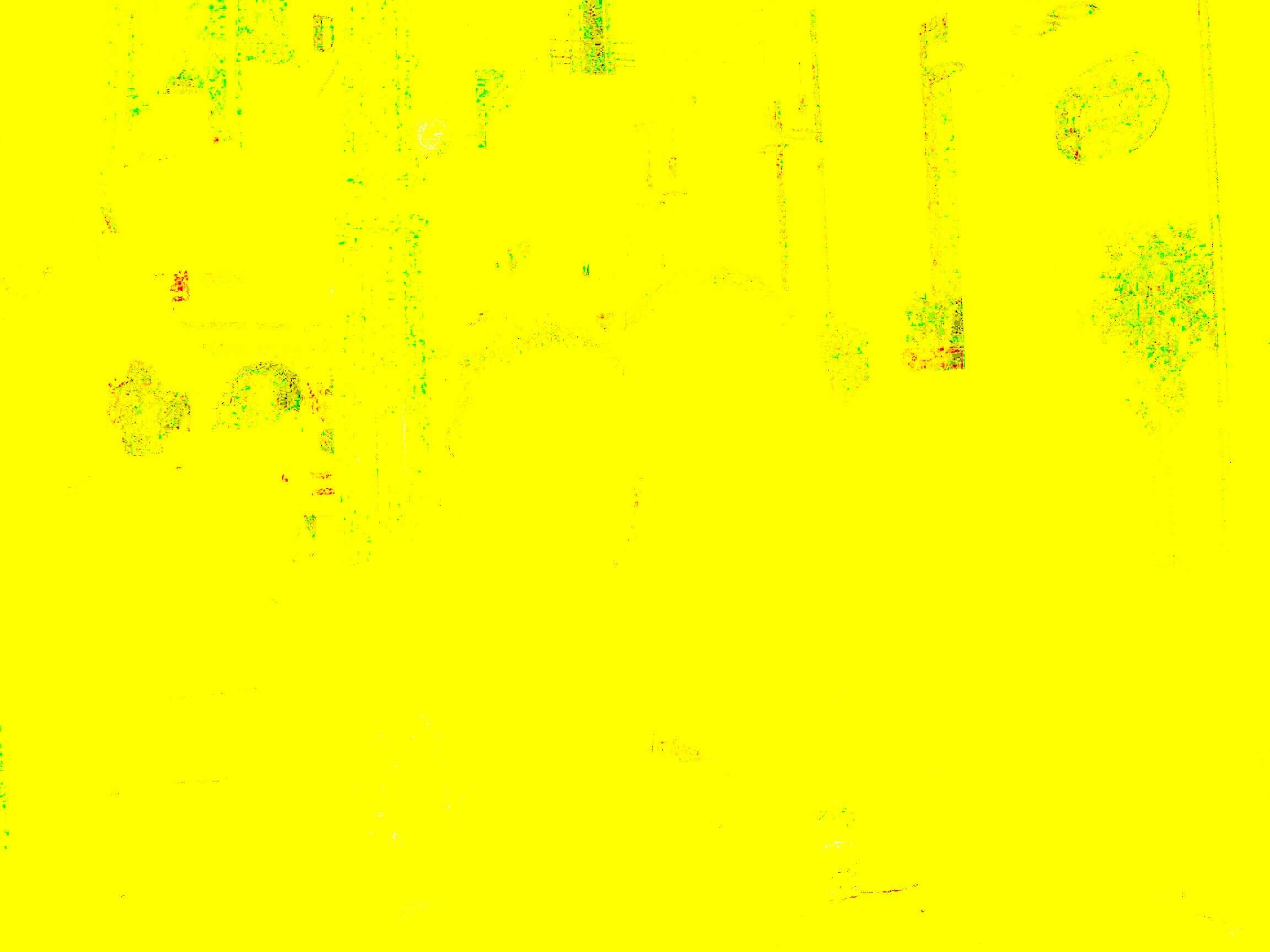}
}
\subfigure{
\includegraphics[width=1.9cm]{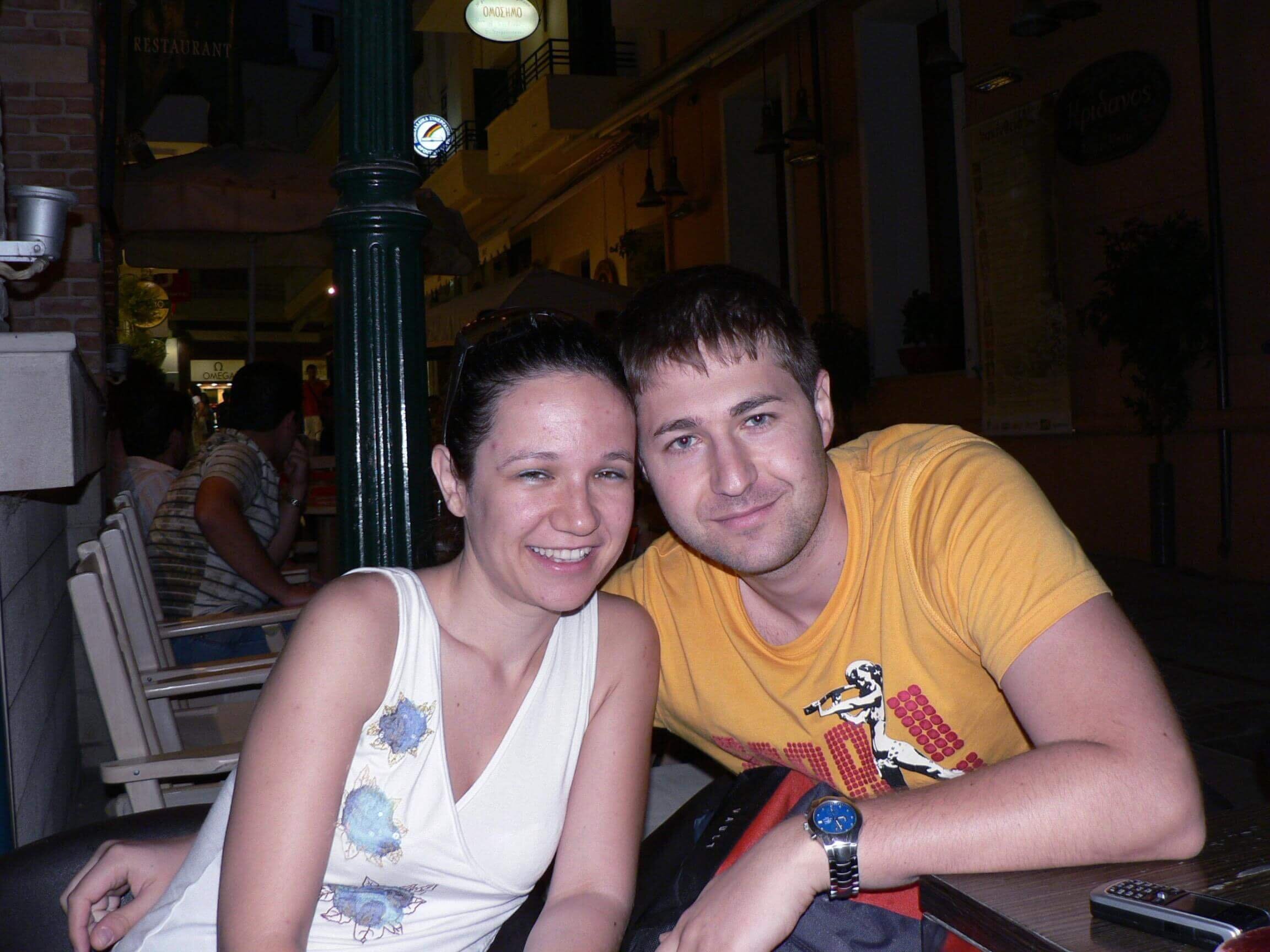}
}
\subfigure{
\includegraphics[width=1.9cm]{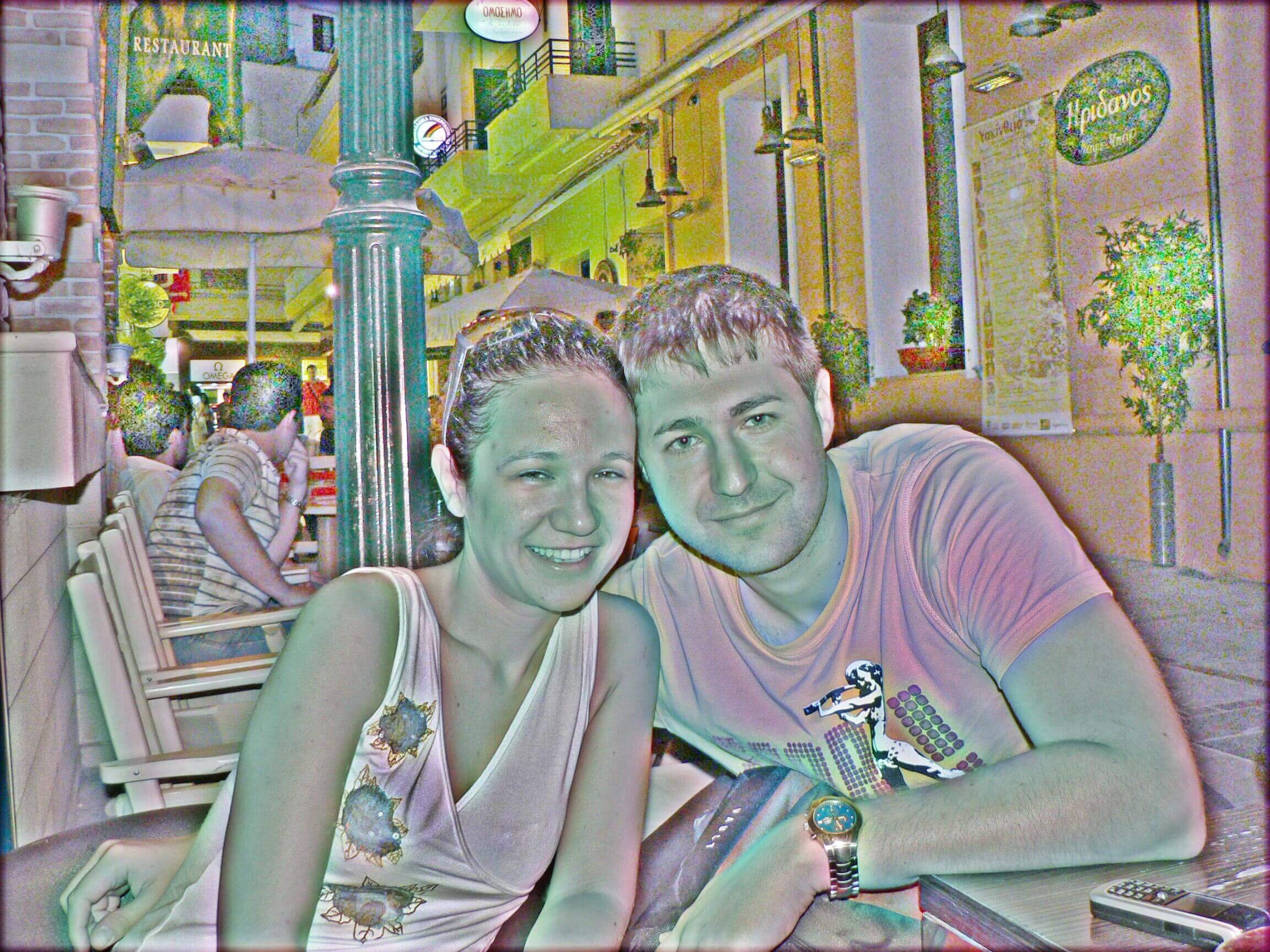}
}
\subfigure{
\includegraphics[width=1.9cm]{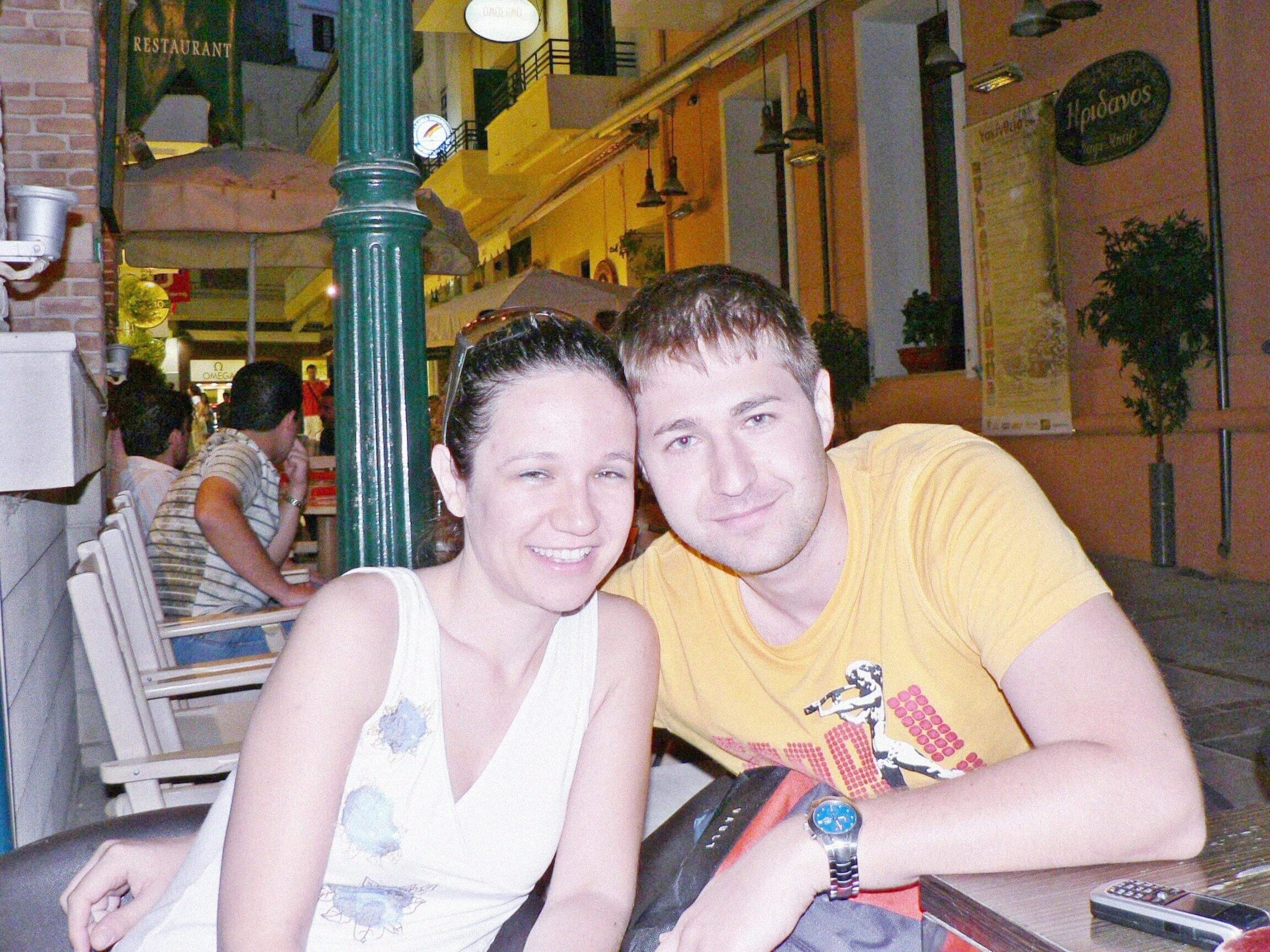}
}
\subfigure{
\includegraphics[width=1.9cm]{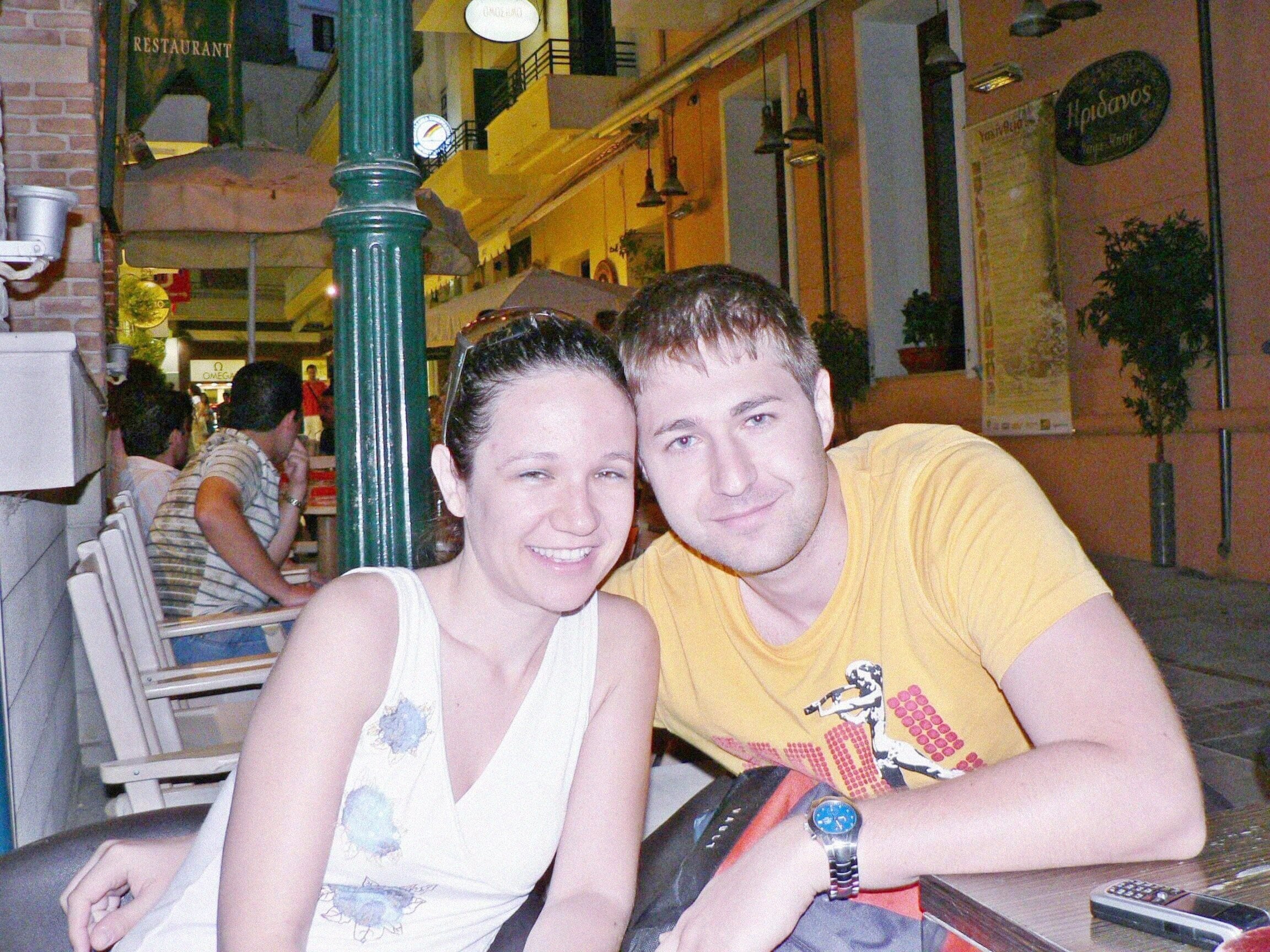}
}
\subfigure{
\includegraphics[width=1.9cm]{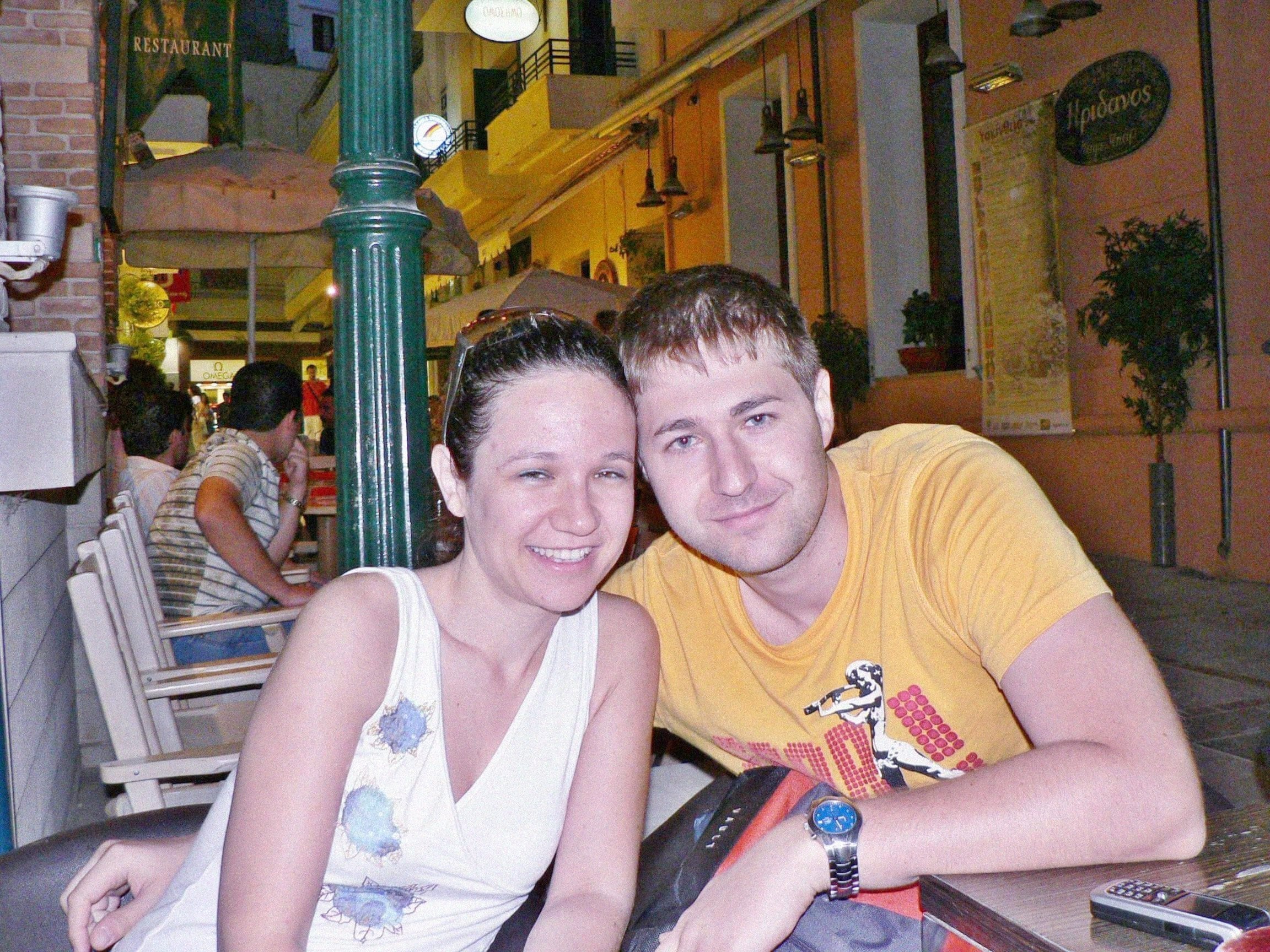}
}
\subfigure{
\includegraphics[width=1.9cm]{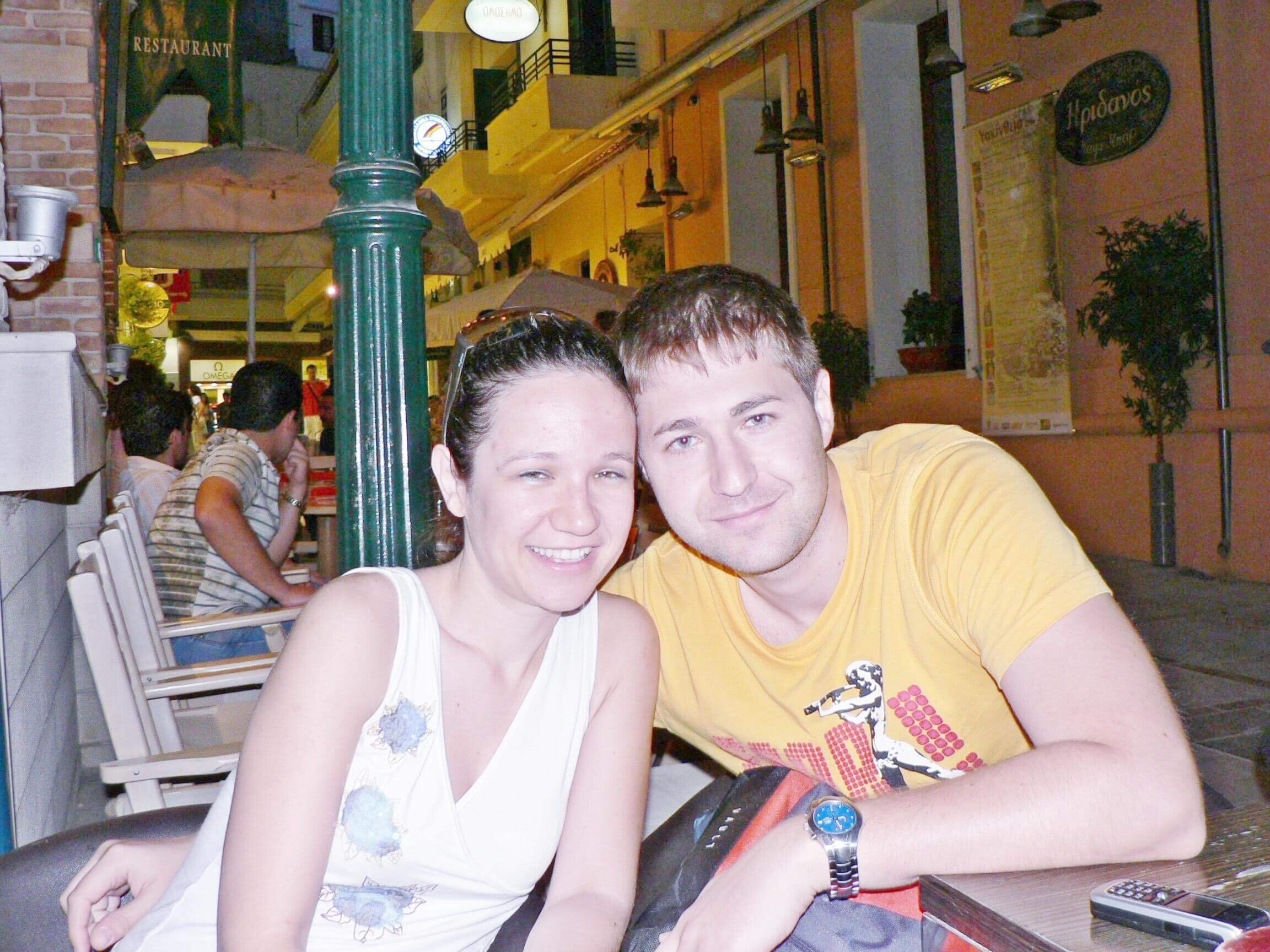}
}

\subfigure[Dark]{
\includegraphics[width=1.9cm]{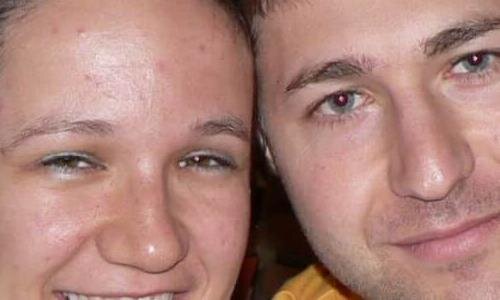}
}
\subfigure[$w/o$ $L_{rgb}$]{
\includegraphics[width=1.9cm]{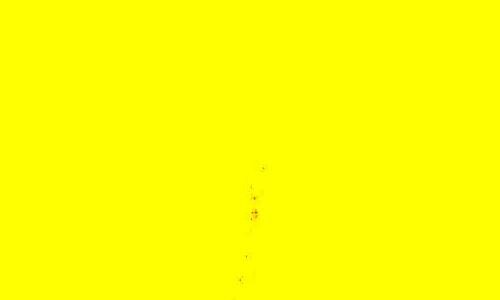}
}
\subfigure[$w/o$ $L_{bri}$] {
\includegraphics[width=1.9cm]{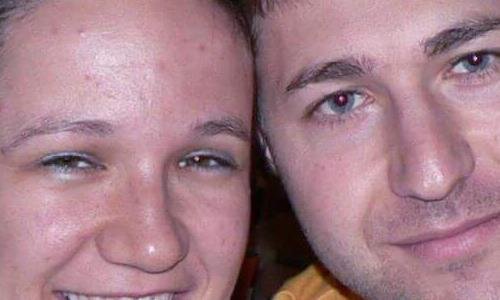}
}
\subfigure[$w/o$ $L_{t v}$]{
\includegraphics[width=1.9cm]{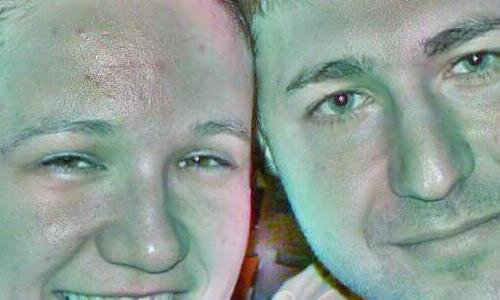}
}
\subfigure[$w/o$ $L_{s p a}$]{
\includegraphics[width=1.9cm]{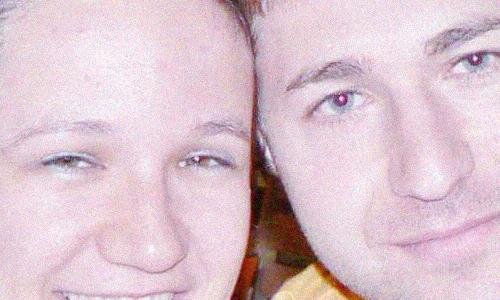}
}
\subfigure[$w/o$ $L_{sem}$]{
\includegraphics[width=1.9cm]{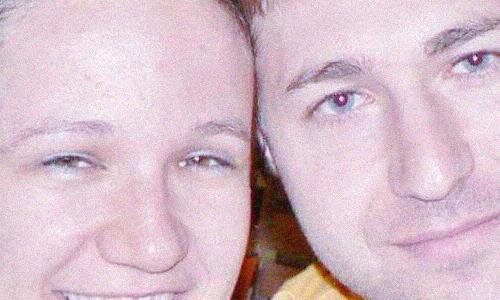}
}
\subfigure[$w/o$ $L_{spa}$ $L_{sem}$]{
\includegraphics[width=1.9cm]{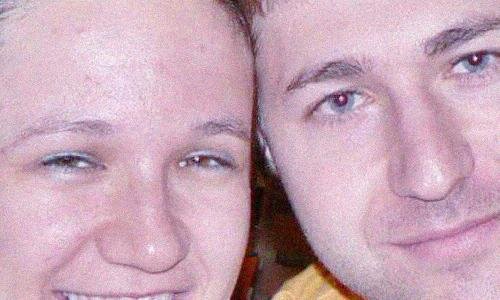}
}
\subfigure[Ours]{
\includegraphics[width=1.9cm]{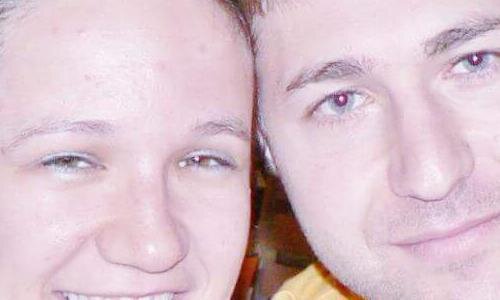}
}
\caption{Visual comparison on loss function ablations. Top row: original enhanced images. Bottom row: cropped enhanced images}
\label{Ablation2}
\end{figure*}

\subsection{Ablation Study}

\begin{table}[t]
\centering
\begin{tabular}{l|l l l}
\hline
Method   & PSNR & SSIM & MSE \\ \hline
w/o $L_{rgb}$ & 8.840 
     &   0.523 
                        &  0.233 

                       \\ 
w/o $L_{bri}$ &  13.06 
    &    0.473 
                       &  0.108 

                        \\
w/o $L_{tv}$  & 16.05 
     &  0.678 
                         &  0.045 

                        \\ 
w/o $L_{spa}$ & 20.53 
     &  0.785
                         &   \textbf{0.023} 

                       \\ 
w/o $L_{sem}$ &  20.15 
    &  0.786
                         &  0.025 
                        \\ 
w/o $L_{spa}$  $L_{sem}$ & 19.86 
 &  0.785
 &  0.026

                        \\ \hline
\textbf{Ours} &  \textbf{20.60} 
 
    &  \textbf{0.793} 
                         &  \textbf{0.023 }
                         
                         \\ \hline
\end{tabular}
\caption{Ablations of loss function on LOL dataset in terms of PSNR $\uparrow$, SSIM $\uparrow$ and MSE $\downarrow$.}
\label{ablation1}
\end{table}

\begin{table}[t]
\centering
\begin{tabular}{l|l l l}
\hline
Method   & PSNR & SSIM & MSE \\ \hline
$A$ = 3  & 20.44 
     & 0.789
     &  \textbf{0.023}
                       \\ 
$A$ = 5  & 20.26 
    &  0.787
    & 0.024
                        \\
$E$ = 0.5   & 17.25 
     & 0.694
                         &  0.054

                        \\ 
$E$ = 0.7 & 20.54 
     & 0.762
                         & 0.024

                       \\  \hline
\textbf{Ours} &  \textbf{20.60}
 
    &  \textbf{0.793} 
                         &  \textbf{0.023 }
                         
                         \\ \hline
\end{tabular}
\caption{Ablations of loss function hyperparameters $A$ and $E$ on LOL dataset in terms of PSNR $\uparrow$, SSIM $\uparrow$ and MSE $\downarrow$. The proposed method uses $A$ = 4 and $E$ = 0.6.}
\label{hyperparam}
\end{table}

We conduct ablation studies to investigate individual loss functions, loss function hyperparameters $A$ and $E$, and weight initialization for USS.

Table \ref{ablation1} displays the loss function ablation. It shows that $L_{bri}$,  $L_{rgb}$ and $L_{tv}$ has large influences on image enhancement results, whereas $L_{spa}$ and $L_{sem}$ have smaller impacts. An additional visual comparison is made on the VV dataset in Fig. \ref{Ablation2}. We find that Model without $L_{rgb}$/$L_{bri}$/$L_{tv}$ have severe color deviation, poor low-light region enhancement, and unnatural artifacts, respectively. We also note that Model without $L_{spa}$ or $L_{sem}$ generates noise-corrupted facial details. Model without $L_{spa}$ and $L_{sem}$ additionally result in poor regional contrast and deficient dark area illumination. 

The ablation of loss function hyperparameters is in Table. \ref{hyperparam}. It can be seen that the proposed value for $A$ and $E$ generates the best result. In short, all loss functions with concurrent settings are essential to reach a promising performance.

The ablation of weight initialization is in Table \ref{AblationWeight}. Although USS pretrained on VOC has slightly better result at LOL, it's performance at the large-scale DarkCityScape is much worse than USS with Gaussian initialization. This phenomenon could result from data bias. Based on this evidence, we conclude that Gaussian initialization is sufficient for a promising outcome.

\begin{table}[t]
\centering
\begin{tabular}{l|l l}
\hline
Weight  & LOL & DarkCityScape \\ \hline
Gaussian & 20.60 / 0.79
     &    \textbf{25.97 / 0.97}
 \\ 
VOC \cite{everingham2010pascal} &  \textbf{20.63 / 0.81}
                &    24.86 / 0.95

                        \\ \hline
\end{tabular}
\caption{Ablations of USS top-down pathway weight initialization on LOL and DarkCityScape dataset in terms of PSNR $\uparrow$/SSIM $\uparrow$.}
\label{AblationWeight}
\end{table}

\begin{table*}[htbp]
\centering
\small
\begin{tabular}{l| l l l l l l |l}
\hline
Method & NPE\cite{wang2013naturalness} & LIME\cite{guo2016lime} & MEF\cite{ma2015perceptual} & DICM\cite{lee2012contrast} & VV & DarkBDD & Average \\ \hline
Dark     & 0.793 / 19.81 
    &  \textbf{0.826} / 21.81 
  
    &  0.738 / 23.56 
  
   & \color{blue}{0.795} / \color{black}{\textbf{21.57} }
 
    & 0.826 / \textbf{23.62} 
 
   &  0.799 / 61.62 
   
   &  0.796 / 28.67
        \\
PIE\cite{fu2015probabilistic}     & 0.801 / 21.72 
 
    &  0.791 / 22.72 
 
    &  0.752 / \textbf{11.02 }
 
   & 0.791 / \color{blue}{21.72} 
 
     &  0.832 / 26.54 
 
  & 0.796 / 53.22 
  
  &  0.794  / \textbf{26.16}
 
        \\
LIME\cite{guo2016lime}    & 0.786 / 18.24 
 
    &  0.774 / \color{blue}{20.44 }
 
    &  0.722 / 15.25 
 
   & 0.758 / 23.48 
 
     & 0.820 / 27.14 
 
   & -/-   
   
   & -/-  
   \\
Retinex\cite{wei2018deep}    & \textbf{0.828} / \color{blue}{16.04}
 
    &   0.794 / 31.47 
 
   &  0.755 / 20.08 
 
   &   0.770 / 29.53 
 
   &  0.824 / 29.58 
 
  &  0.792 / \color{blue}{50.77}
  
  &  0.794 / 29.57
 
       \\
MBLLEN\cite{lv2018mbllen}   &  0.793 / 34.46 
 
   & 0.768 / 30.26 
 
     &  0.717 / 37.44 
 
   & 0.787 / 32.44 

     & 0.719 / 26.13 
 
   &  0.772 / 51.40 
 
 &  0.759 / 35.35
       \\
KinD\cite{zhang2019kindling}    & 0.792 / 19.65 
 
    &  0.766 / 39.29 
 
    &  0.747 / 31.36 
 
   &  0.776 / 32.71 
 
   &  0.814 / 29.34 
 
  & 0.778 / \textbf{49.38} 
 
 &  0.779 / 33.62
        \\ 
Zero-DCE\cite{guo2020zero}  & \color{blue}{0.814} / \color{black}{17.06 }
 
    &  \color{blue}{0.811} / \color{black}{21.40 }
 
    &  \color{blue}{0.762} / \color{black}{16.84 }
 
   & 0.777 / 27.35 
 
     & \color{blue}{0.835} / \color{blue}{24.26} 
 
   & \color{blue}{0.800} / \color{black}{59.37 }
 
 &  \color{blue}{0.800} /  \color{black}{27.71}
        \\ \hline
\textbf{Ours} & 0.786 / \textbf{13.25 }
 
    &  0.807 / \textbf{19.99 }
 
    &  \textbf{0.785} / \color{blue}{13.92 }
 
   &  \textbf{0.801} / 26.12 
 
    &  \textbf{0.836} / 31.72 
 
  &\textbf{0.815} / 57.06 
 
 & \textbf{0.805} / \color{blue}{27.01}
        \\ \hline
\end{tabular}
\caption{UNIQUE $\uparrow$ / BRISQUE $\downarrow$ Comparison on NPE, LIME, MEF, DICM, VV and DarkBDD}
\label{niqe}
\end{table*}

\begin{table*}[htbp]
\small
\renewcommand\tabcolsep{1.8pt}
\centering
\begin{tabular}{l|llllll|l}
\hline
Dataset & Dark & PIE\cite{fu2015probabilistic} &  Retinex\cite{wei2018deep} & MBLLEN\cite{lv2018mbllen} & KinD\cite{zhang2019kindling} & Zero-DCE\cite{guo2020zero} & \textbf{Ours} \\ \hline
LOL      &  13.20/0.48/0.106 
    &  20.18/0.77/0.025
  &  17.59/0.54/0.044 
  &  \textbf{21.21}/\textbf{0.84}/\textbf{0.016} 
    &  19.29/0.76/0.040 
     &  20.38/0.78/\color{blue}{0.023} 
     & \color{blue}{20.60}/\color{blue}{0.79}/\color{blue}{0.023} 
  \\ 
DCS    &  16.22/0.77/0.026 
    &  17.49/0.83/0.020  
  &  10.54/0.65/0.091 
  &  22.52/0.88/0.007 
    &  12.28/0.73/0.062 
     &  \color{blue}{22.59}/\color{blue}{0.94}/\color{blue}{0.006}
     & \textbf{25.97}/\textbf{0.97}/\textbf{0.004}
  \\
  \hline
\end{tabular}
\caption{PSNR $\uparrow$ / SSIM $\uparrow$ / MSE $\downarrow$ Comparison on LOL and DarkCityScape (DCS)}
\label{PSNR}
\end{table*}

\subsection{Model Comparisons}

\noindent
\textbf{Quantitative Comparison}
We conduct quantitative comparisons for different models. Traditional methods PIE \cite{fu2015probabilistic} and LIME \cite{guo2016lime} were excluded for efficiency comparison because they unfit GPU acceleration. For all tables, we use \textbf{bold} for the best score and \textcolor{blue}{blue} for the second-best score. ‘-’ indicates that a result is unavailable due to excessive image size for a particular model.


Table \ref{niqe} shows the comparison on NPE, LIME, MEF, DICM, and VV datasets. Our model has the best average UNIQUE and the second best average BRISQUE. Table \ref{PSNR} shows the model comparison on LOL and DarkCityScape dataset. Our method is the second best in LOL and is the best in the more challenging extreme low-light dataset DarkCityScape.


Table \ref{effi} shows that the proposed model is computationally the most efficient. The proposed model's run time is 0.001 second for a single image (i.e., 1000 images can be processed within 1 second). Besides, the significantly fewer FLOPs indicates our model fits low-light video enhancement. Furthermore, the proposed method is ideal for mobile devices due to the small parameters. 


\begin{table}[t]
\centering
\small
\begin{tabular}{l | l l l l}
\hline
Method & RT$\downarrow$ & Params$\downarrow$ & FLOPs$\downarrow$ & Score$\uparrow$ \\ \hline
Retinex\cite{wei2018deep}      &  0.121   &  0.555         & 587.5   &  2.30        \\ 
MBLLEN\cite{lv2018mbllen}          & 0.526    &  0.450          & 301.1   &  3.05        \\ 
KinD\cite{zhang2019kindling}        &  0.147   &  8.160         &  575.0      & \color{blue}{3.36}         \\ 
EnlightenGAN\cite{jiang2021enlightengan}  & 0.008    & 8.637          & 273.2     &   2.94
    \\ 
Zero-DCE\cite{guo2020zero}& \color{blue}{0.003 }    &  \color{blue}{0.079 }         &  \color{blue}{84.99}   &  2.60        \\ \hline
\textbf{Ours}            & \textbf{0.001}   &  \textbf{0.011}   & \textbf{0.120}    & \textbf{4.04}         \\ \hline
\end{tabular}
\caption{Model Efficiency and User Study Score Comparison. Images of size $1200\times900$ are selected for experiments. `RT' is the inference time in seconds per image. `Params' are the numbers of trainable parameters in millions per image, and `FLOPs' are the numbers of floating-point operations in billions per image. All model inference is conducted with a single Nvidia GeForce RTX 2080 Ti GPU.}
\label{effi}
\end{table}

\begin{figure*}[htbp] 
\centering
\includegraphics[width=3.5cm]{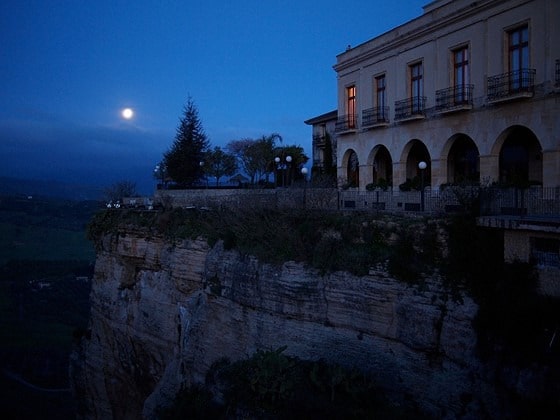}
\includegraphics[width=3.5cm]{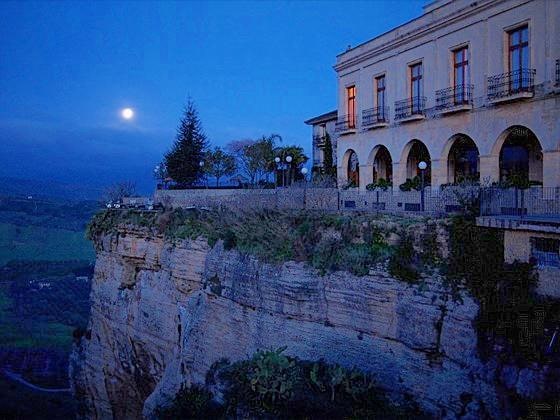}
\includegraphics[width=3.5cm]{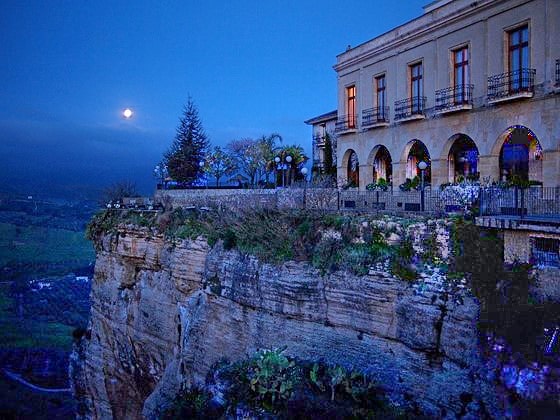}
\includegraphics[width=3.5cm]{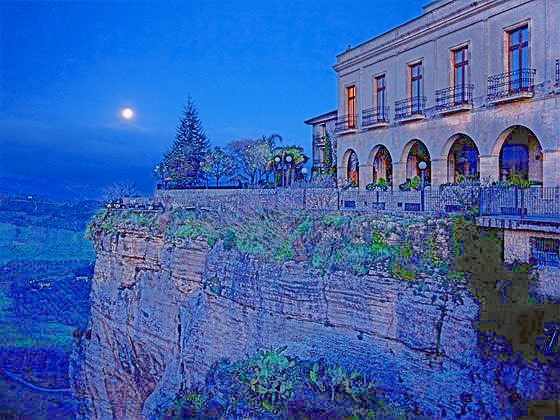}\\
\includegraphics[width=3.5cm]{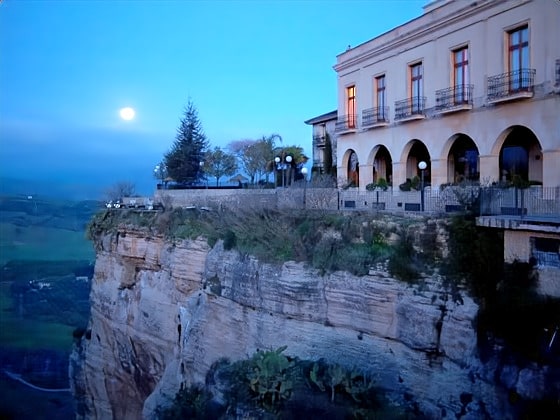}
\includegraphics[width=3.5cm]{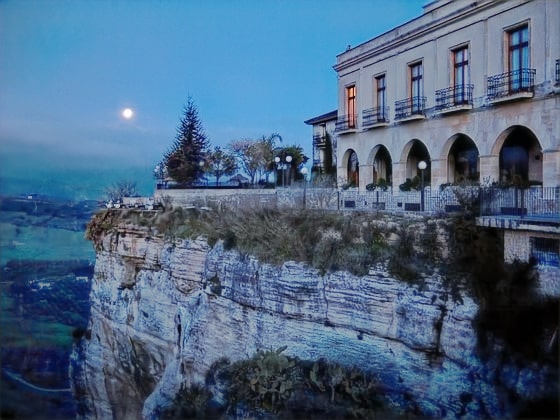}
\includegraphics[width=3.5cm]{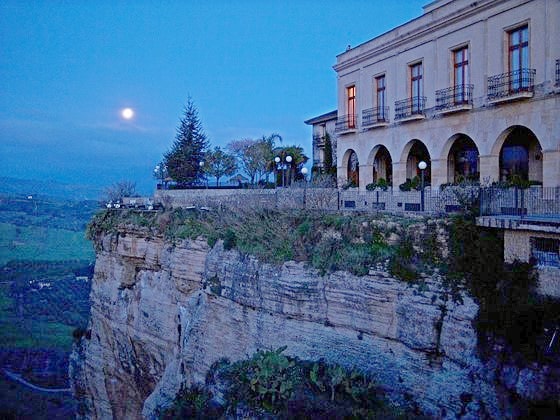}
\includegraphics[width=3.5cm]{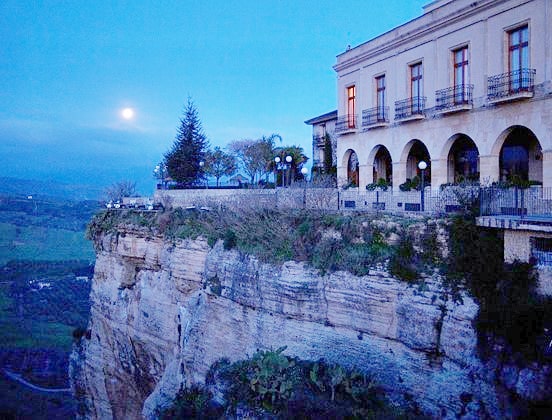} \\


\includegraphics[width=3.5cm]{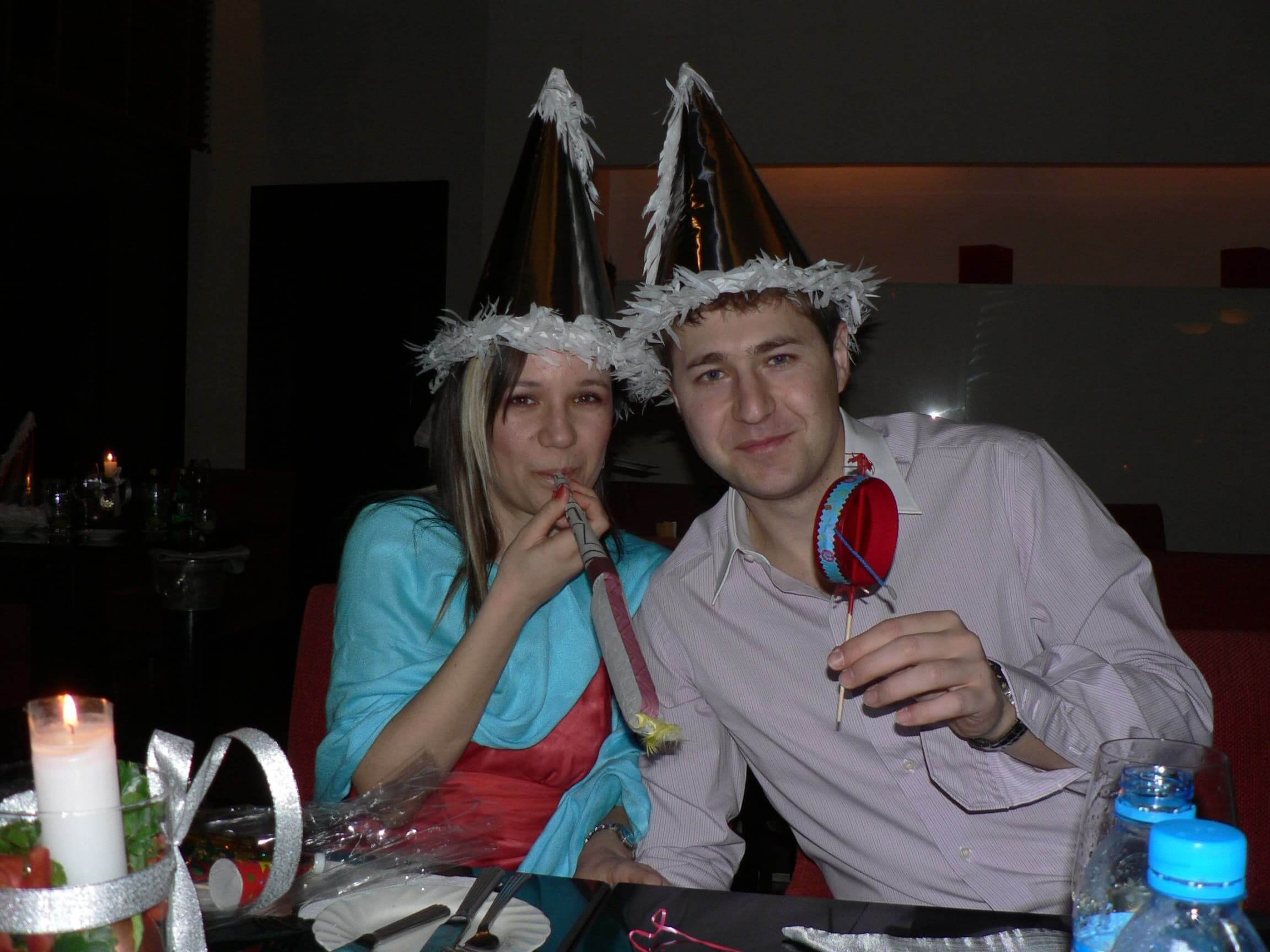}
\includegraphics[width=3.5cm]{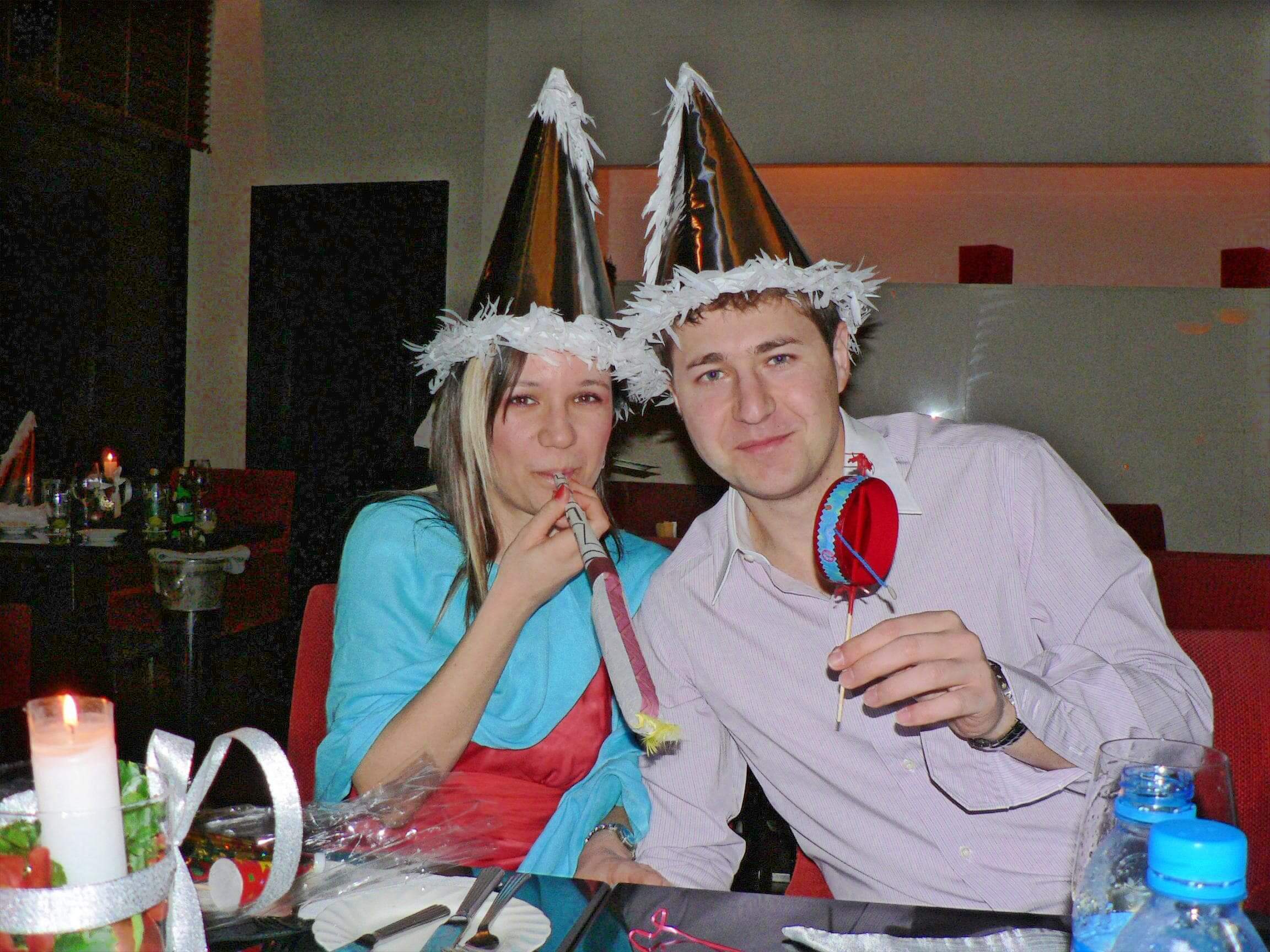}
\includegraphics[width=3.5cm]{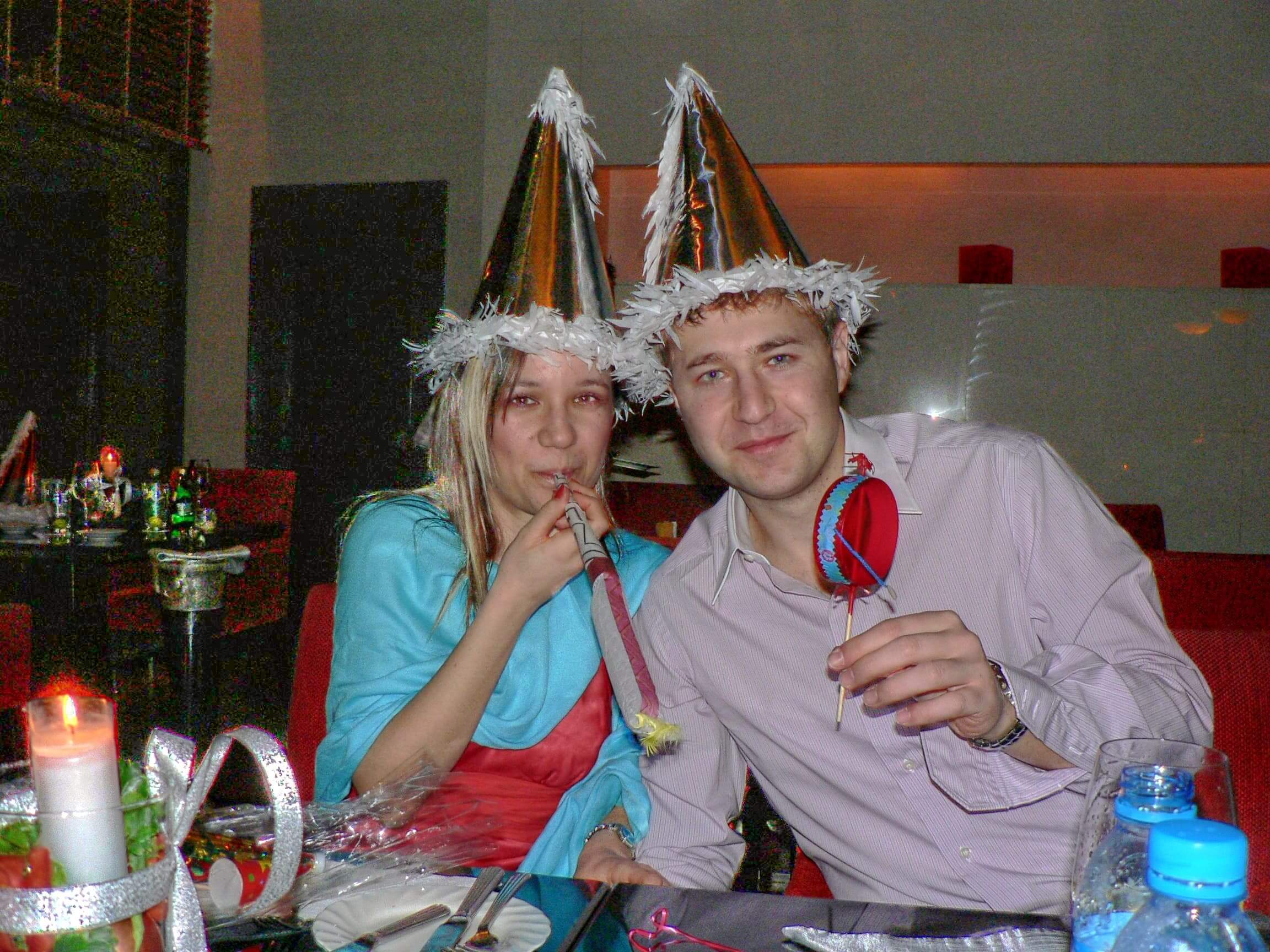}
\includegraphics[width=3.5cm]{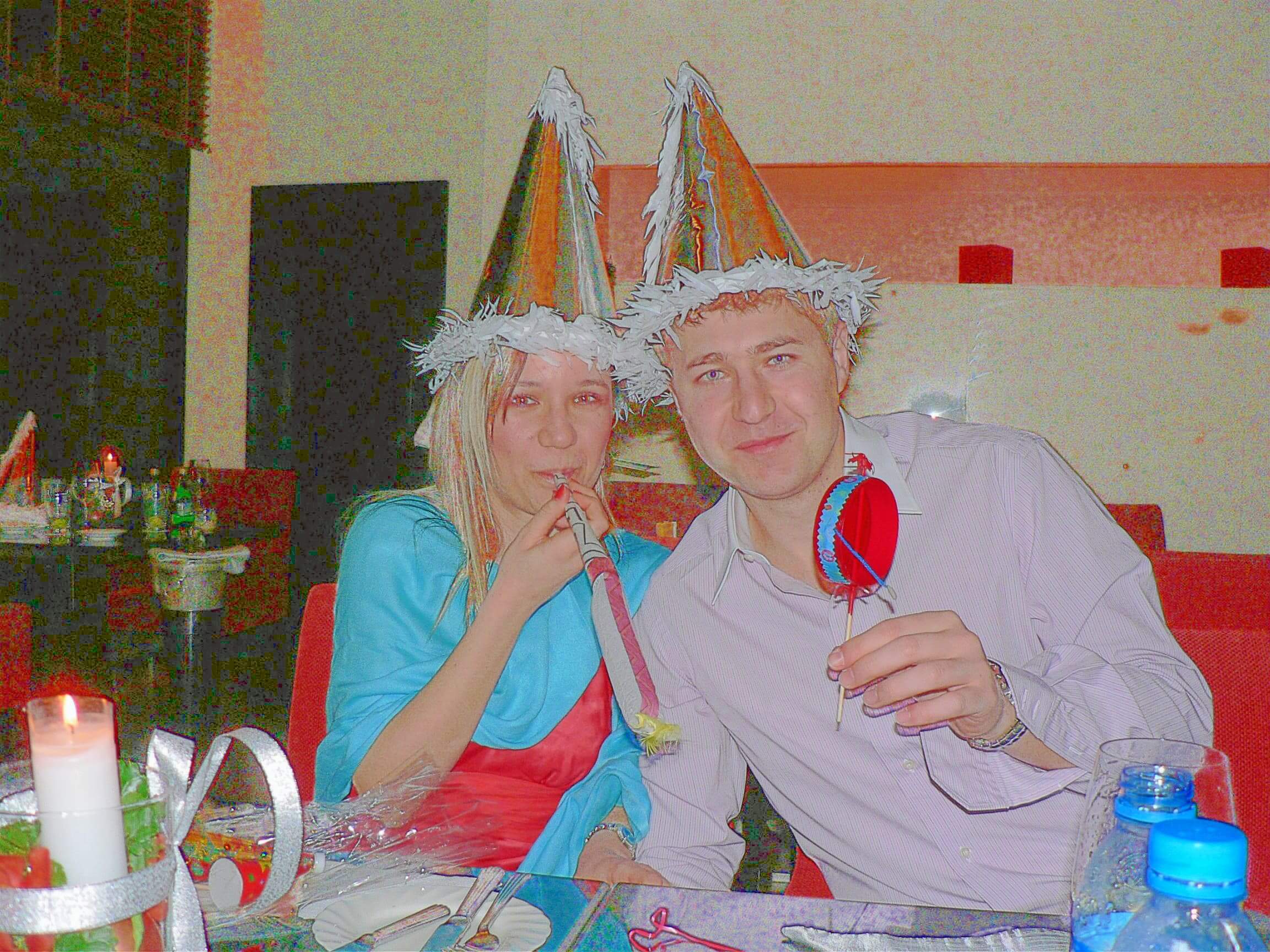}\\
\includegraphics[width=3.5cm]{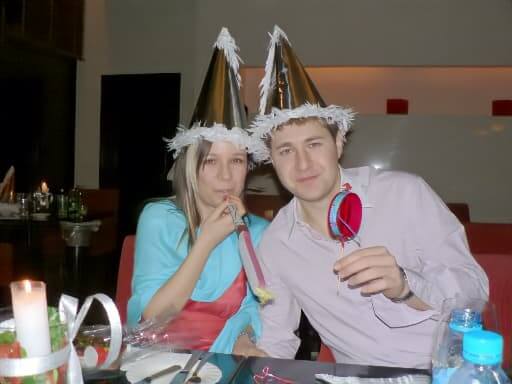}
\includegraphics[width=3.5cm]{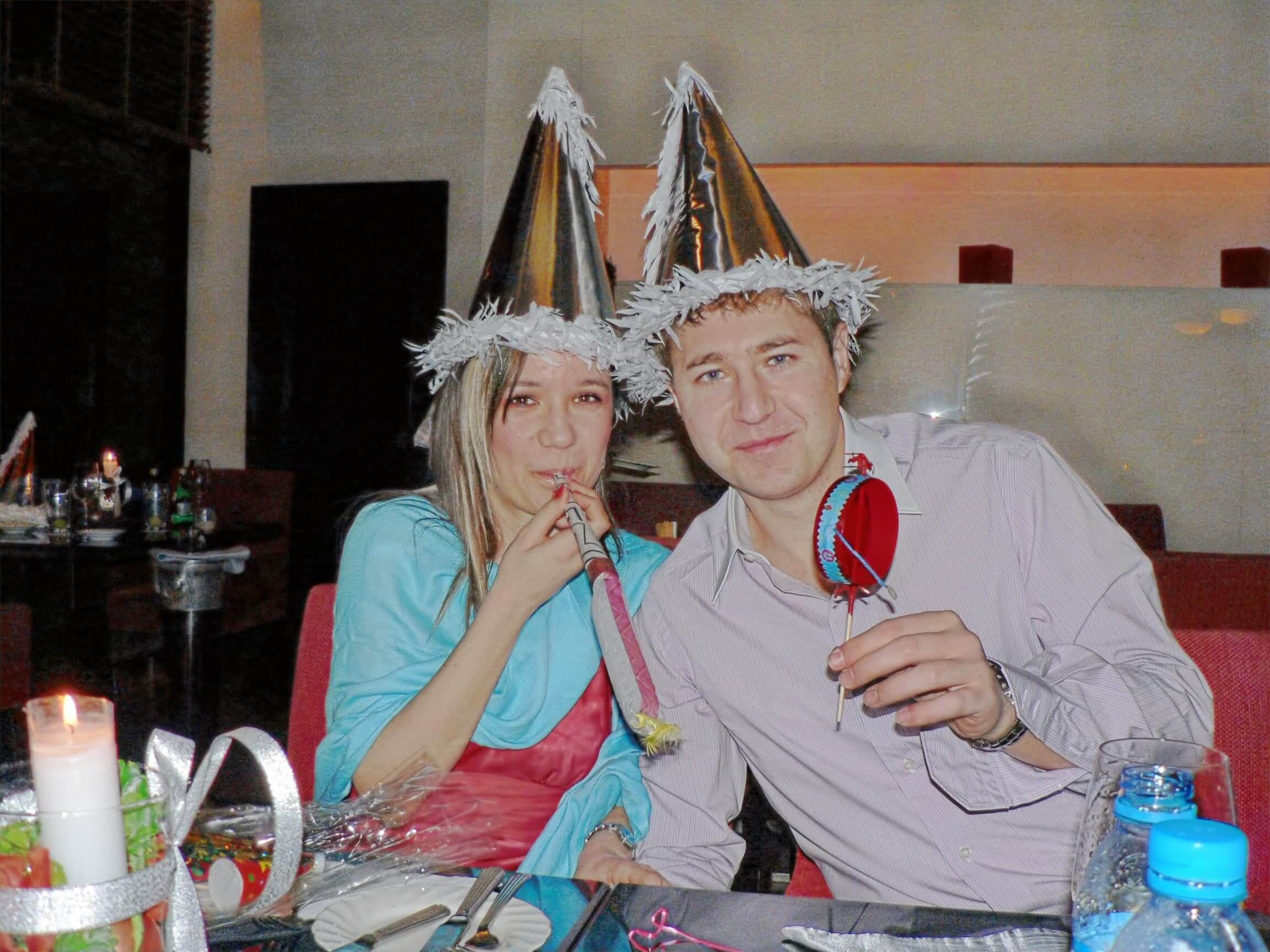}
\includegraphics[width=3.5cm]{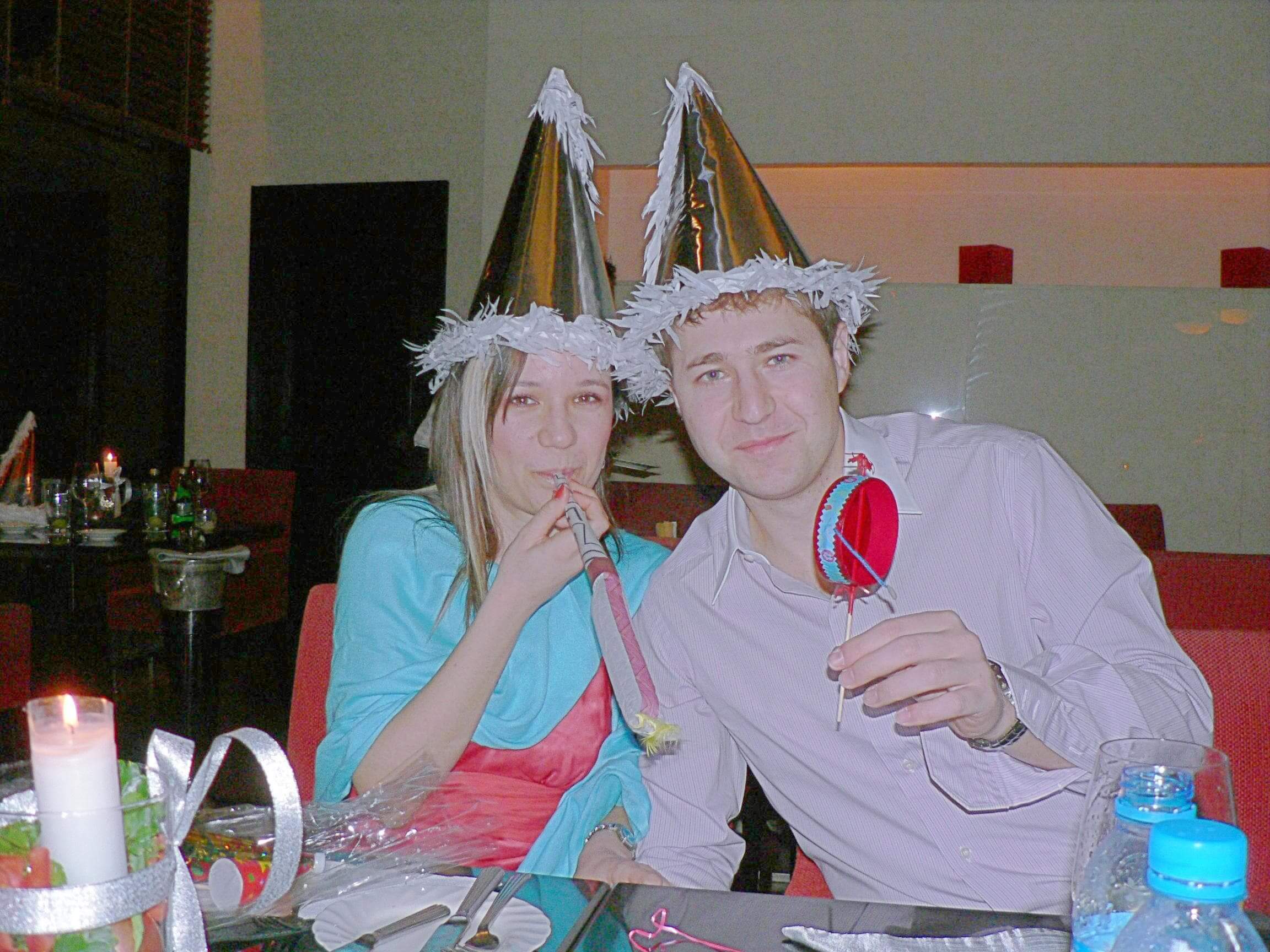}
\includegraphics[width=3.5cm]{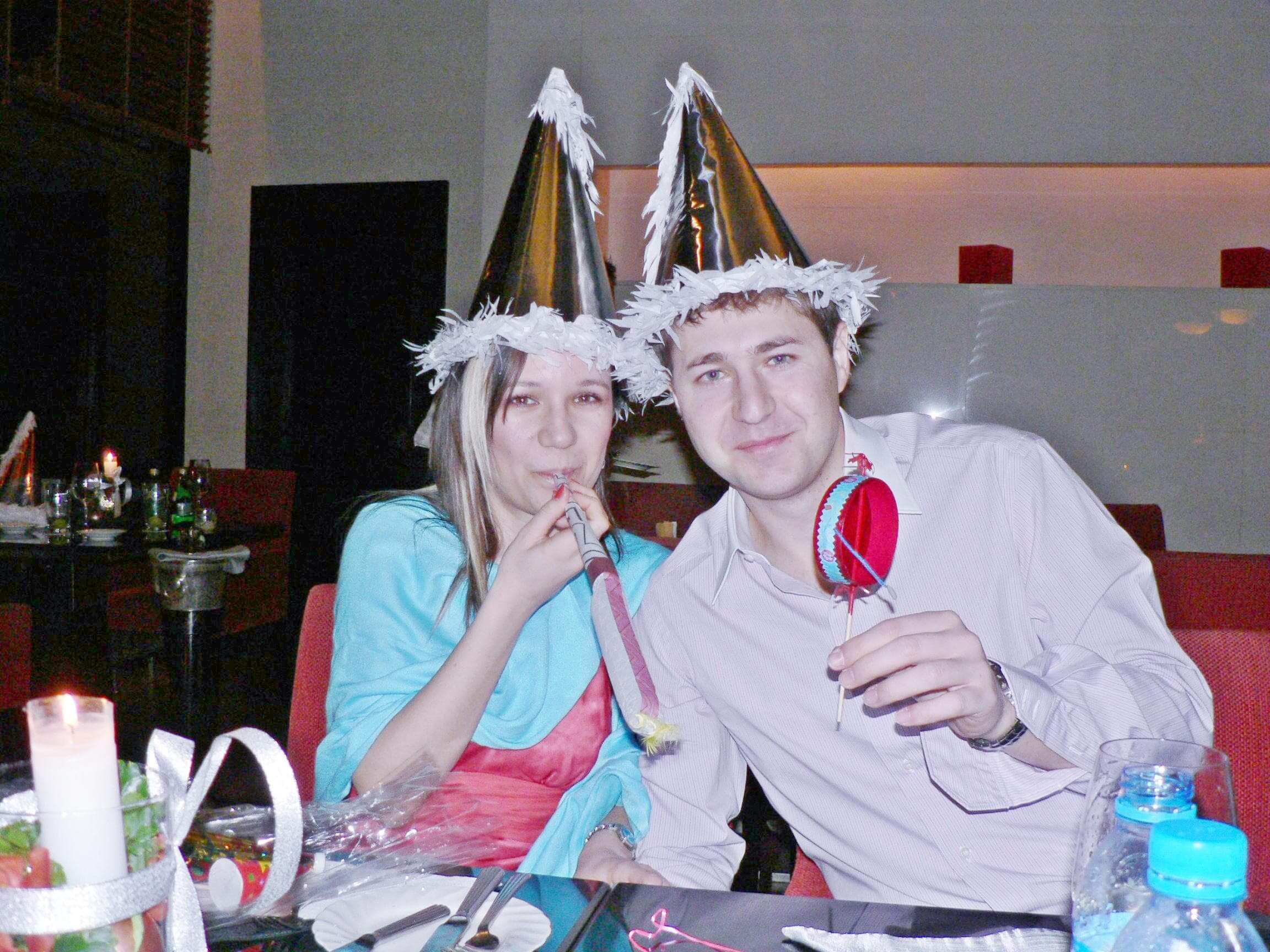}
\caption{Visual Comparison on LIME \cite{guo2016lime} (top two rows) and VV dataset (bottom two rows). For each two rows, from left to right, and from top to bottom: Dark, PIE\cite{fu2015probabilistic}, LIME\cite{guo2016lime}, Retinex\cite{wei2018deep}, MBLLEN\cite{lv2018mbllen}, KinD\cite{zhang2019kindling}, Zero-DCE\cite{guo2020zero}, Ours}
\label{Visual1}
\end{figure*}


\begin{figure*}[htbp] 
\centering
\includegraphics[width=3.5cm]{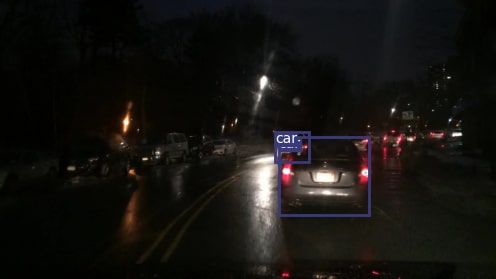}
\includegraphics[width=3.5cm]{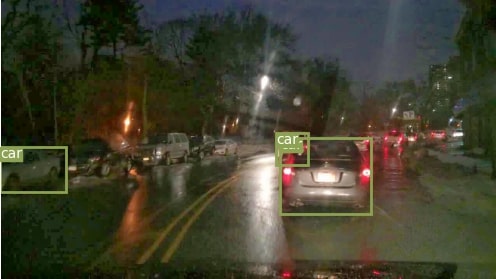}
\includegraphics[width=3.5cm]{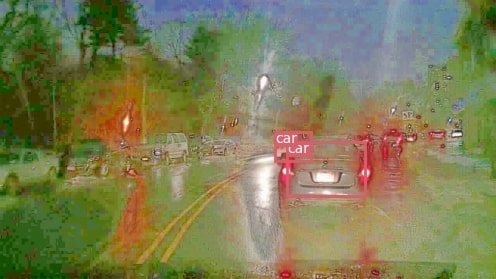}
\includegraphics[width=3.5cm]{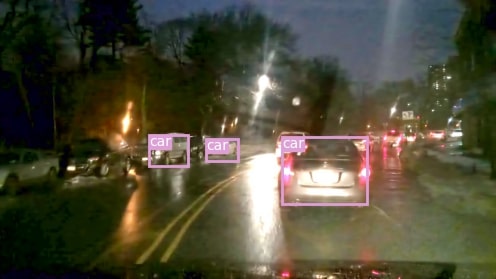} \\
\includegraphics[width=3.5cm]{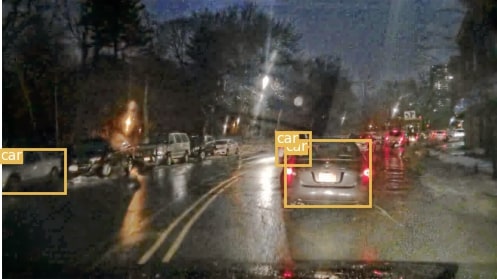}
\includegraphics[width=3.5cm]{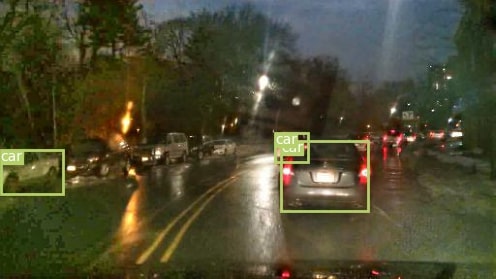}
\includegraphics[width=3.5cm]{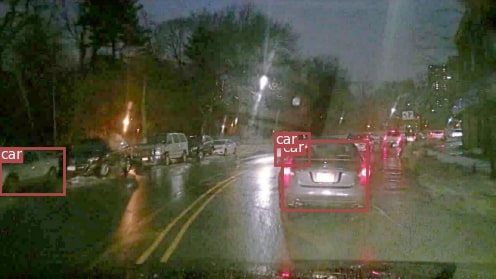}
\includegraphics[width=3.5cm]{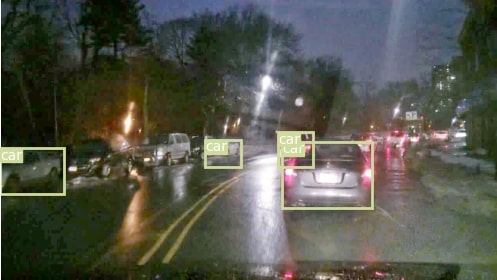}
\caption{Object Detection Results on DarkBDD. From left to right, and from top to bottom: Dark, PIE\cite{fu2015probabilistic}, Retinex\cite{wei2018deep}, MBLLEN\cite{lv2018mbllen}, KinD\cite{zhang2019kindling}, EnlightenGAN\cite{jiang2021enlightengan}, Zero-DCE\cite{guo2020zero}, Ours}
\label{Object}
\end{figure*}

\begin{figure*}[htbp] 
\centering
\includegraphics[width=3.5cm]{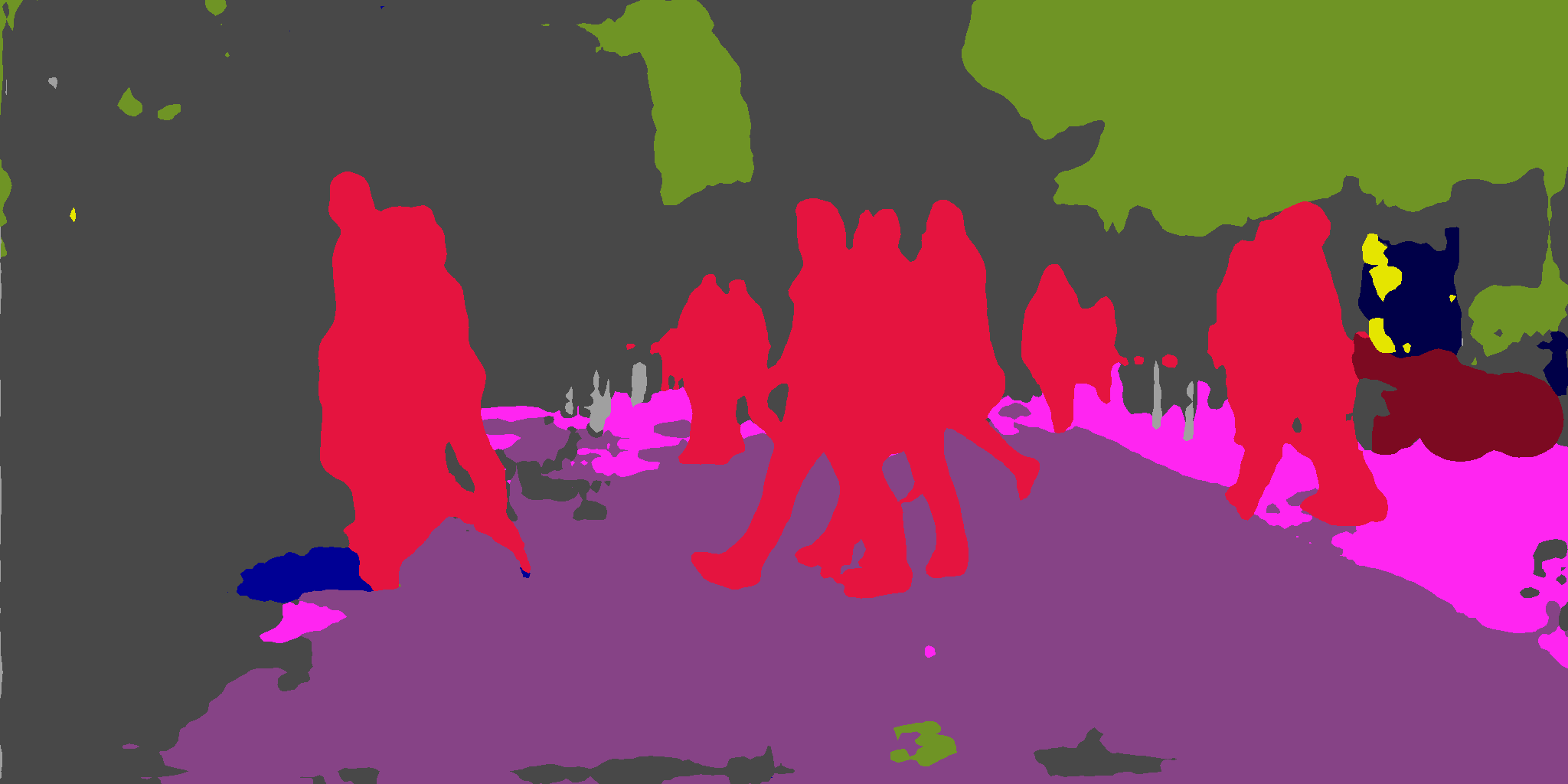}
\includegraphics[width=3.5cm]{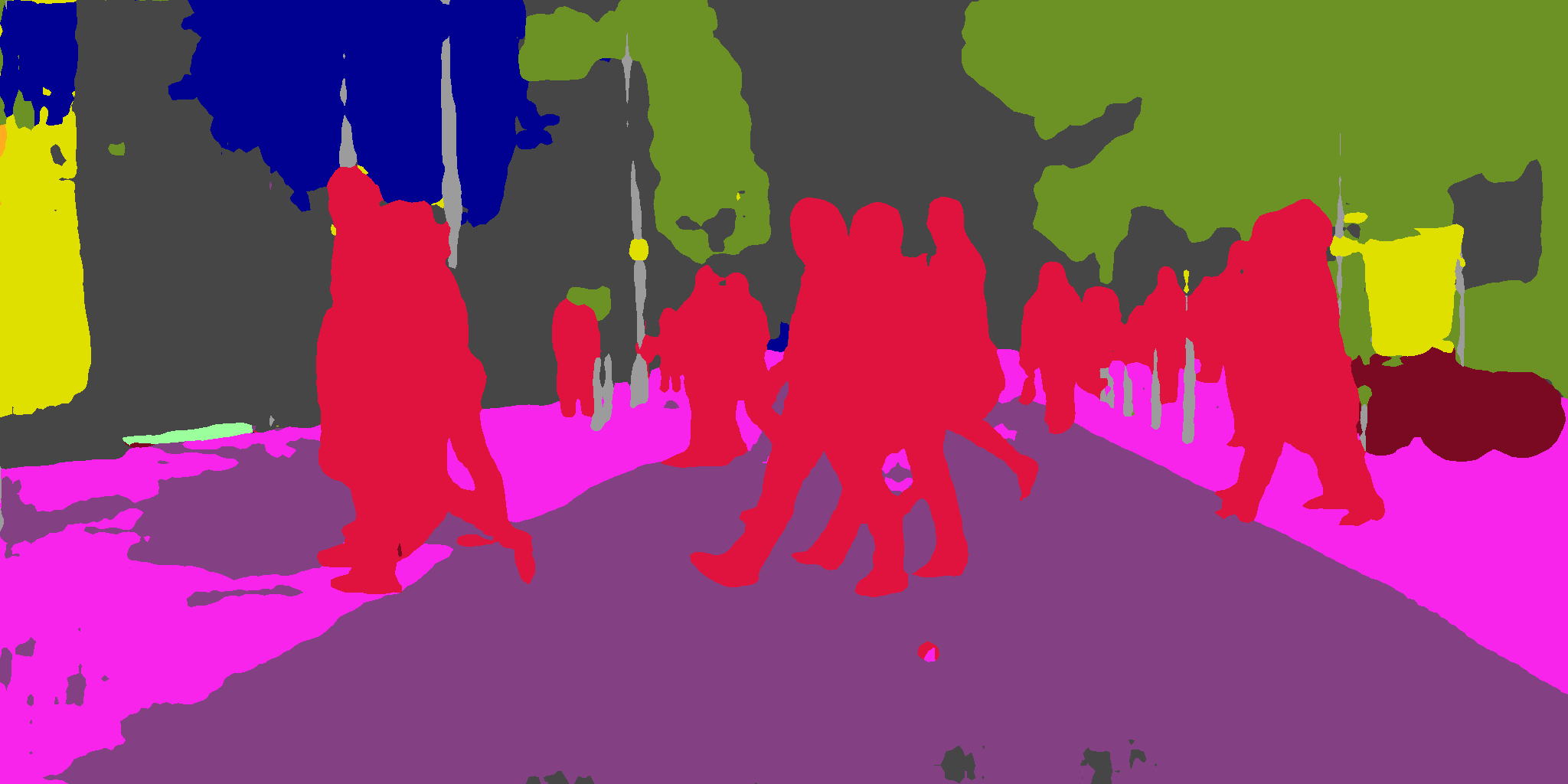}
\includegraphics[width=3.5cm]{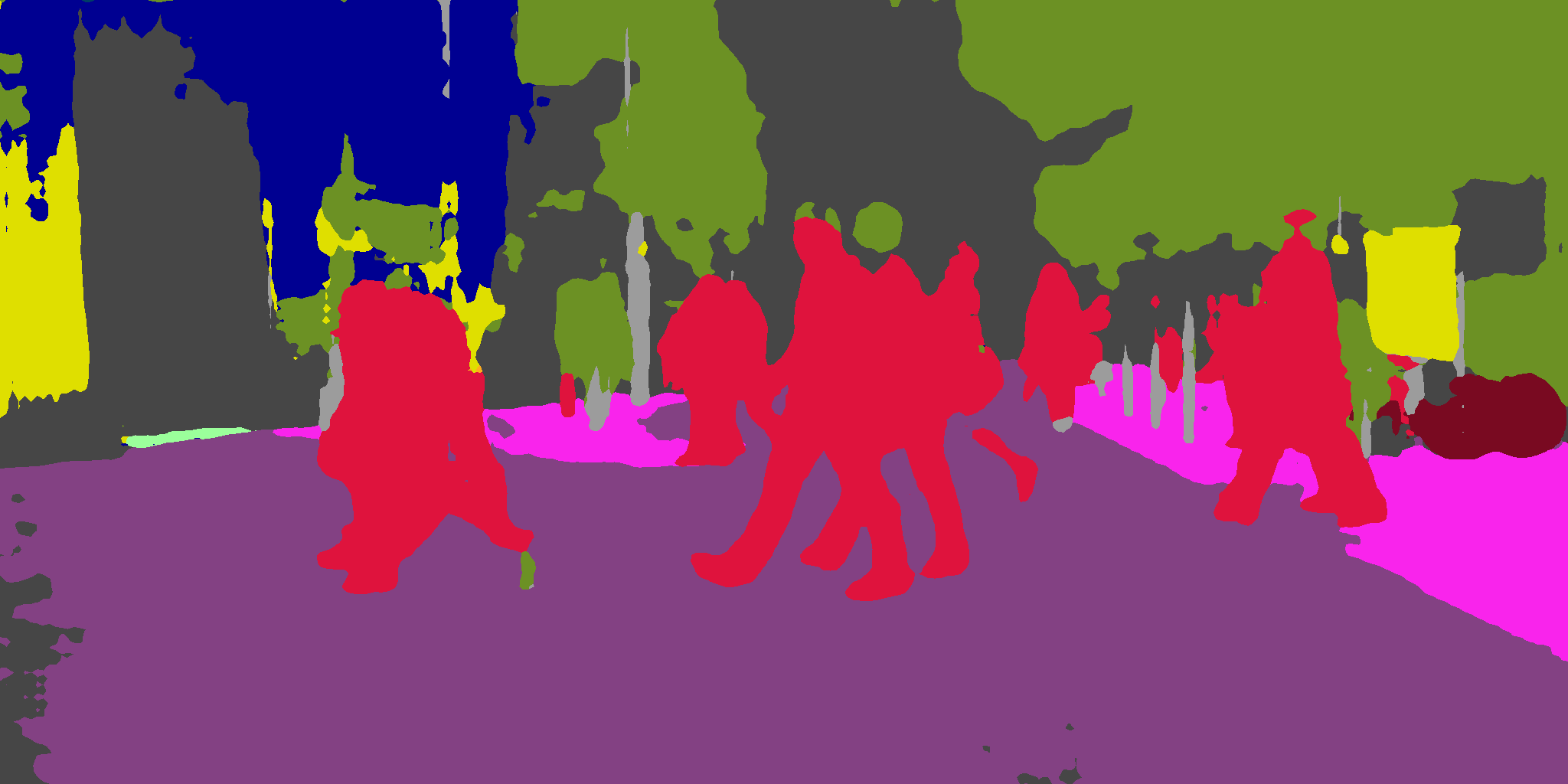}
\includegraphics[width=3.5cm]{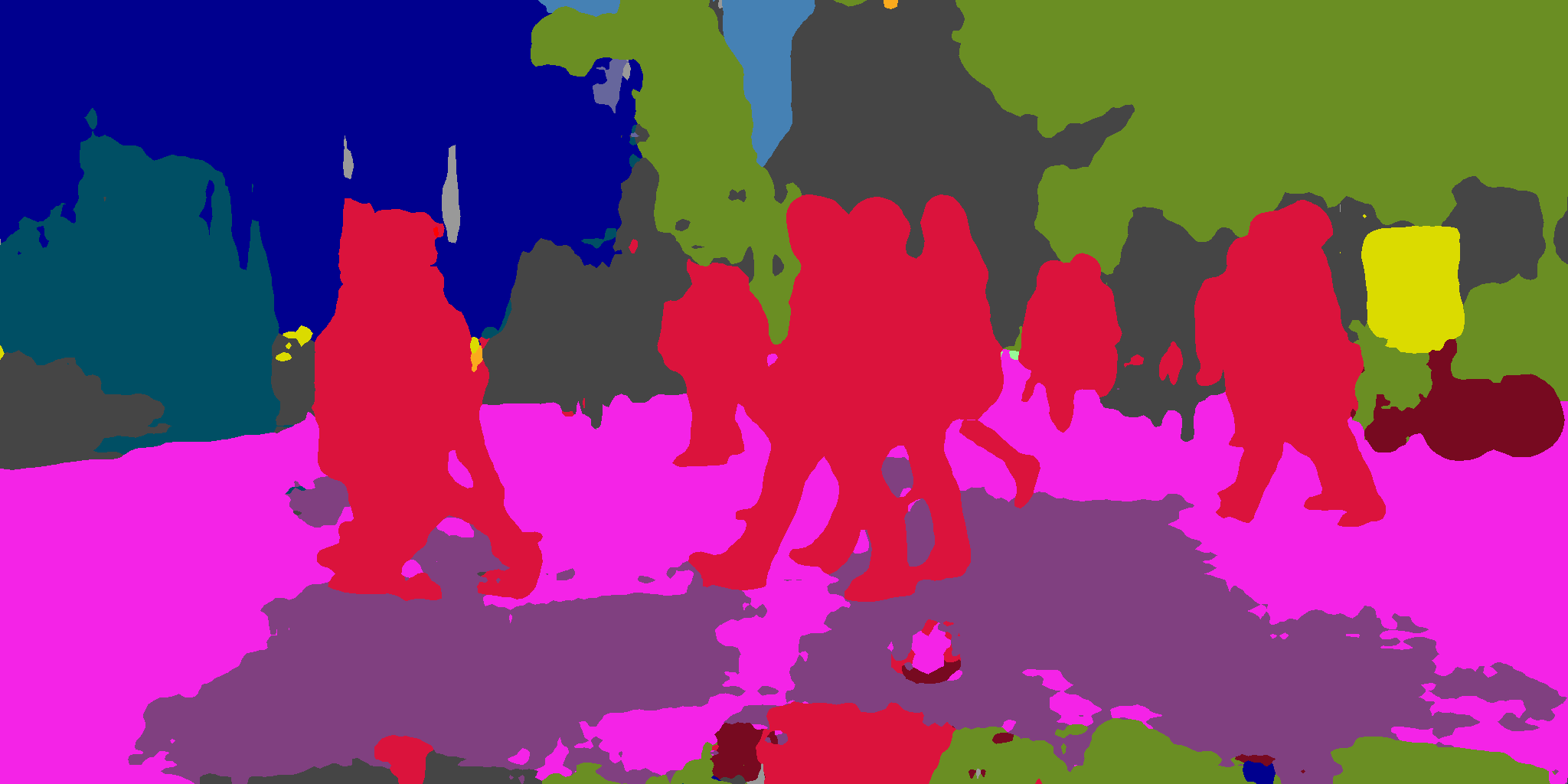}
\includegraphics[width=3.5cm]{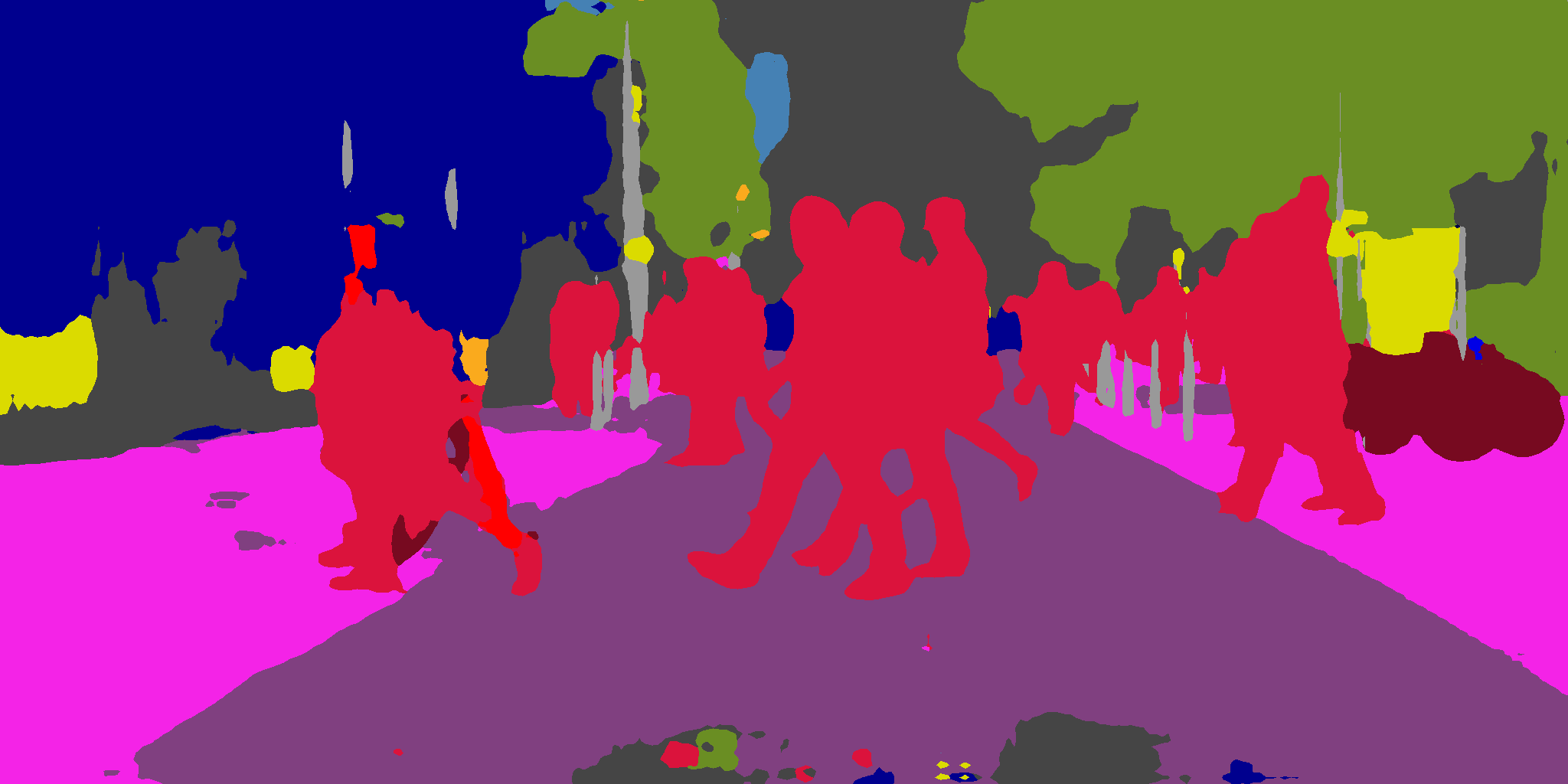}
\includegraphics[width=3.5cm]{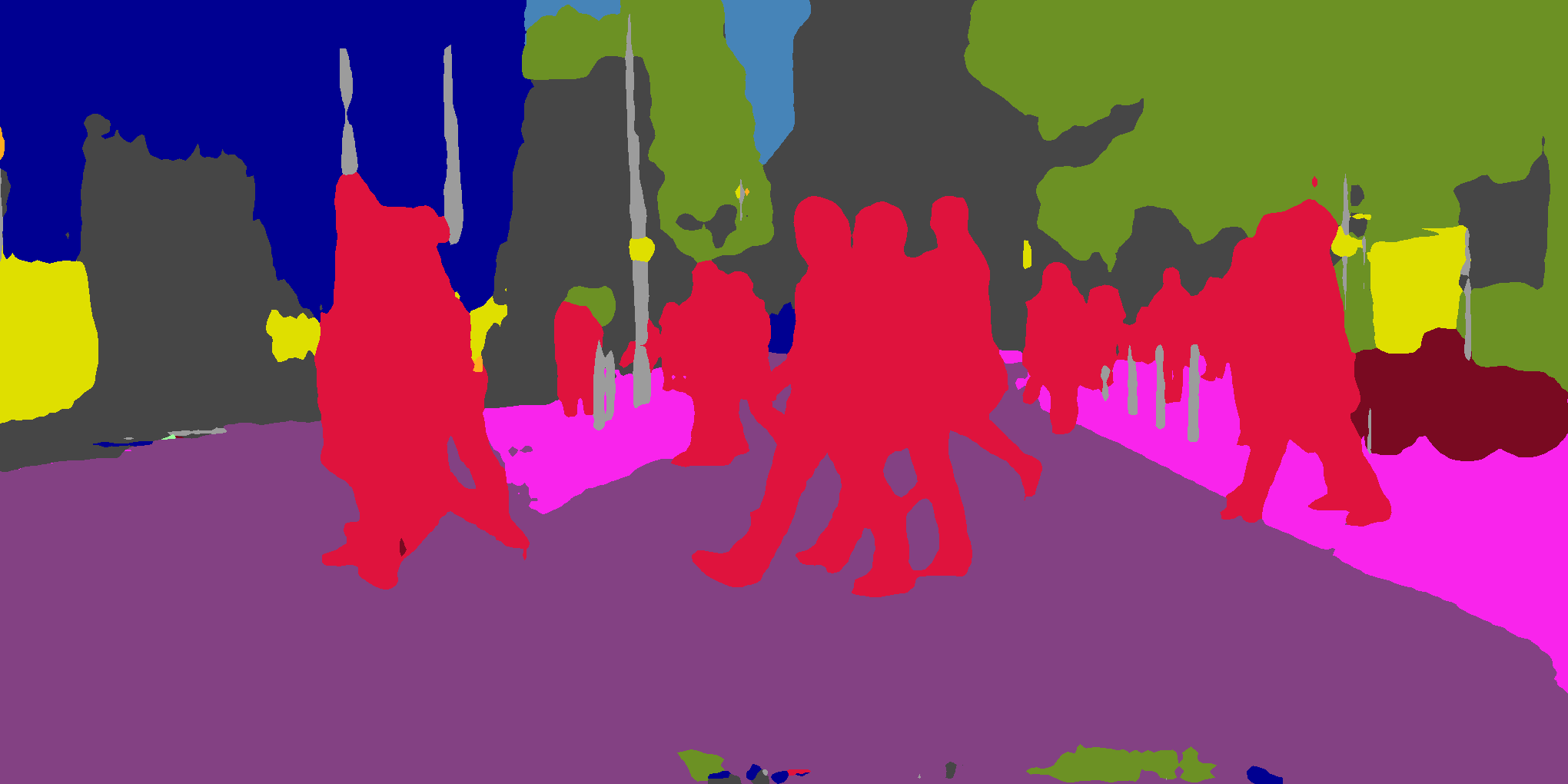}
\includegraphics[width=3.5cm]{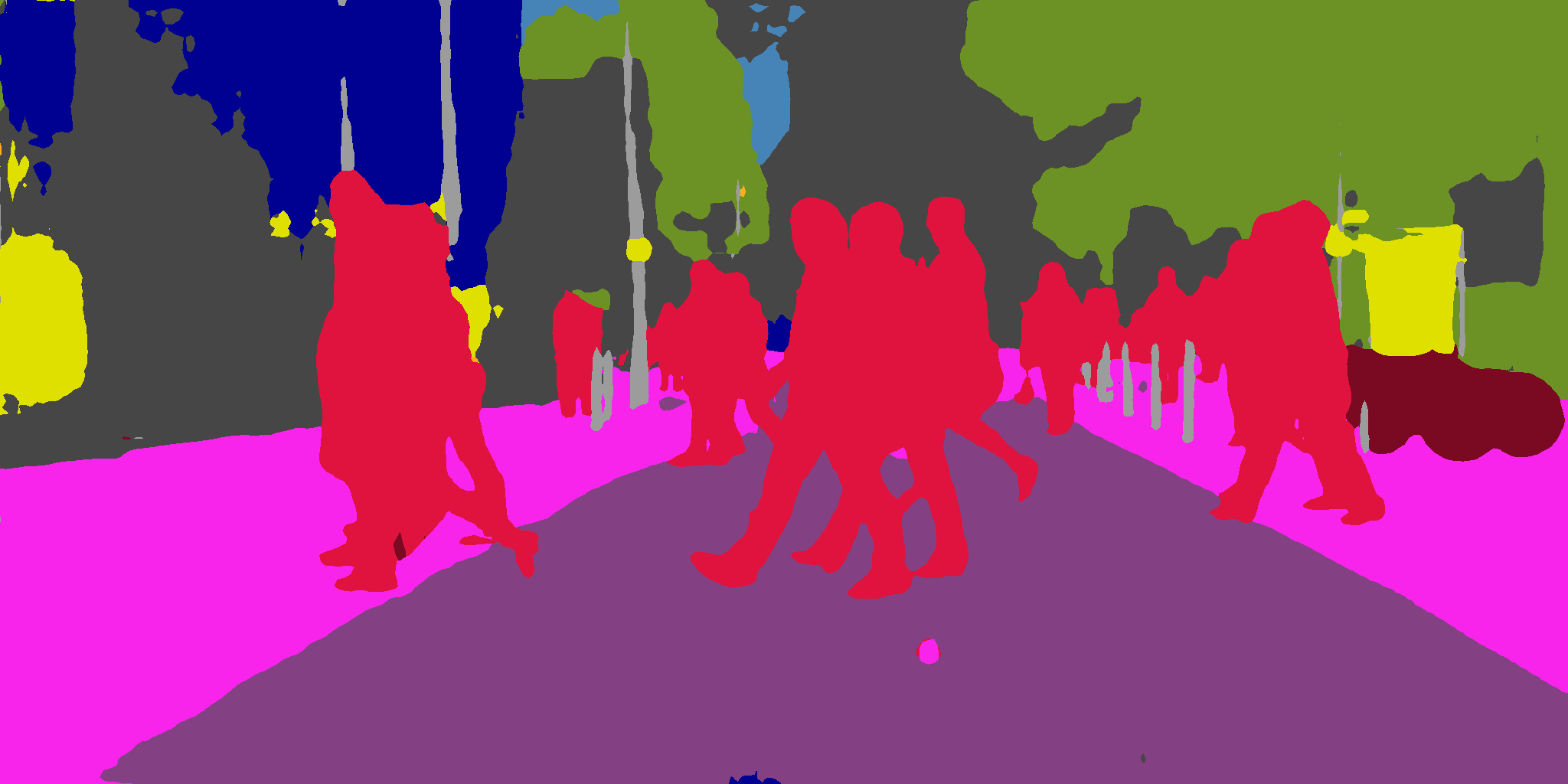}
\includegraphics[width=3.5cm]{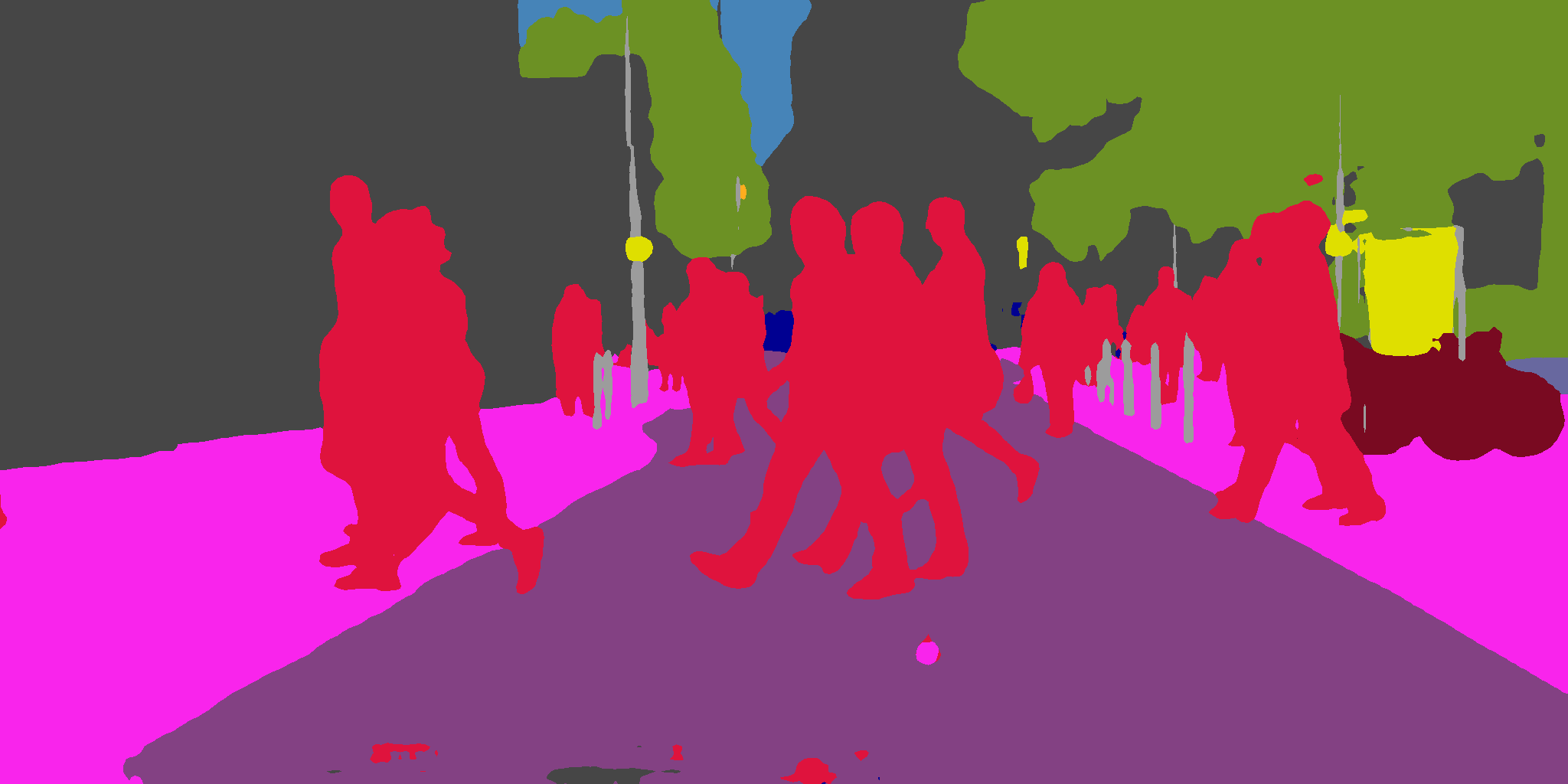}
\caption{Semantic Segmentation Results on DarkCityScape. From left to right, and from top to bottom: Dark, PIE\cite{fu2015probabilistic}, Retinex\cite{wei2018deep}, MBLLEN\cite{lv2018mbllen}, KinD\cite{zhang2019kindling}, Zero-DCE\cite{guo2020zero}, Ours, GroundTruth}
\label{Semantic}
\end{figure*}

\noindent
\textbf{Quantitative Comparison}
We present the visual comparison of different models at Fig. \ref{Visual1}. It can be seen that the proposed model significantly enhances the dark regions, maintains color balance and image contrast, and presents natural exposure with sufficient facial detail.

\subsection{Low-Light Detection and Segmentation}

We utilize the object detection model Yolov3 \cite{redmon2018yolov3} and the semantic segmentation model PSPNet \cite{zhao2017pyramid} to investigate how different low-light image enhancement methods are beneficial to the high-level tasks.

We show the perceptual comparison of object detection in Fig. \ref{Object}. PIE, Retinex, and Zero-DCE improve the image's brightness but meanwhile introduce blur and noise. KinD and EnlightenGAN, though somewhat aids detection, produces unnatural background artifacts. In comparison, our model helps detect the greatest numbers of cars. 

We display the perceptual comparison of semantic segmentation in Fig. \ref{Semantic}. Retinex and MBLLEN leave large areas of incorrect segmentation. PIE, Zero-DCE and KinD's enhancement leads to accurate pedestrians segmentation but undesirable holes in the left sidewalk and background trees. In comparison, the proposed method is the closest to the groundtruth. Finally, we show a quantitative comparison on semantic segmentation using mean Intersection Over Union (mIOU) and mean Pixel Accuracy (mPA) in Table \ref{MIOUCity}. Our model has the best score for both mIOU and mPA.

\begin{table}[t]
\small
\renewcommand\tabcolsep{1.5pt}
\centering
\begin{tabular}{l|llllll|l}
\hline
Metric & Dark & PIE & Retinex & MBLLEN & KinD & Zero-DCE & \textbf{Ours} \\ \hline
mIOU      &  54.49

    &  61.97

  & 57.96
 
  &  51.98

    &  63.42

     & \color{blue}{64.36}
 
     & \textbf{65.87}
 
  \\
mPA    &  70.76

    &  68.89

   & 66.76

  & 59.06

    &  71.69

     & \color{blue}{74.20}

      & \textbf{74.50}

   \\ \hline
\end{tabular}
\caption{mIOU ($\%$) $\uparrow$ and mPA ($\%$) $\uparrow$ Comparison on DarkCityScape}
\label{MIOUCity}
\end{table}

\subsection{Low-Light Video Enhancement}

Unlike prior researches that have been pivotal to single low-light image enhancement, we also examine the performance on a nighttime aerial video. The video is captured using a drone camera with 24 FPS and a resolution of $960\times540$. The video is 41 seconds long and is saved as MP4. We conduct a user study to quantitatively assess the enhancement performance. Specifically, we ask 50 adult participants to rate the enhancement result (video) of five models, including EnlightenGAN \cite{jiang2021enlightengan}, KinD \cite{zhang2019kindling}, Retinex \cite{wei2018deep}, Zero-DCE \cite{guo2020zero} and our model, and we report the result in Table \ref{effi}. More video enhancement results are in the supplementary material.

\begin{figure}[H]
\centering
\includegraphics[width=2cm]{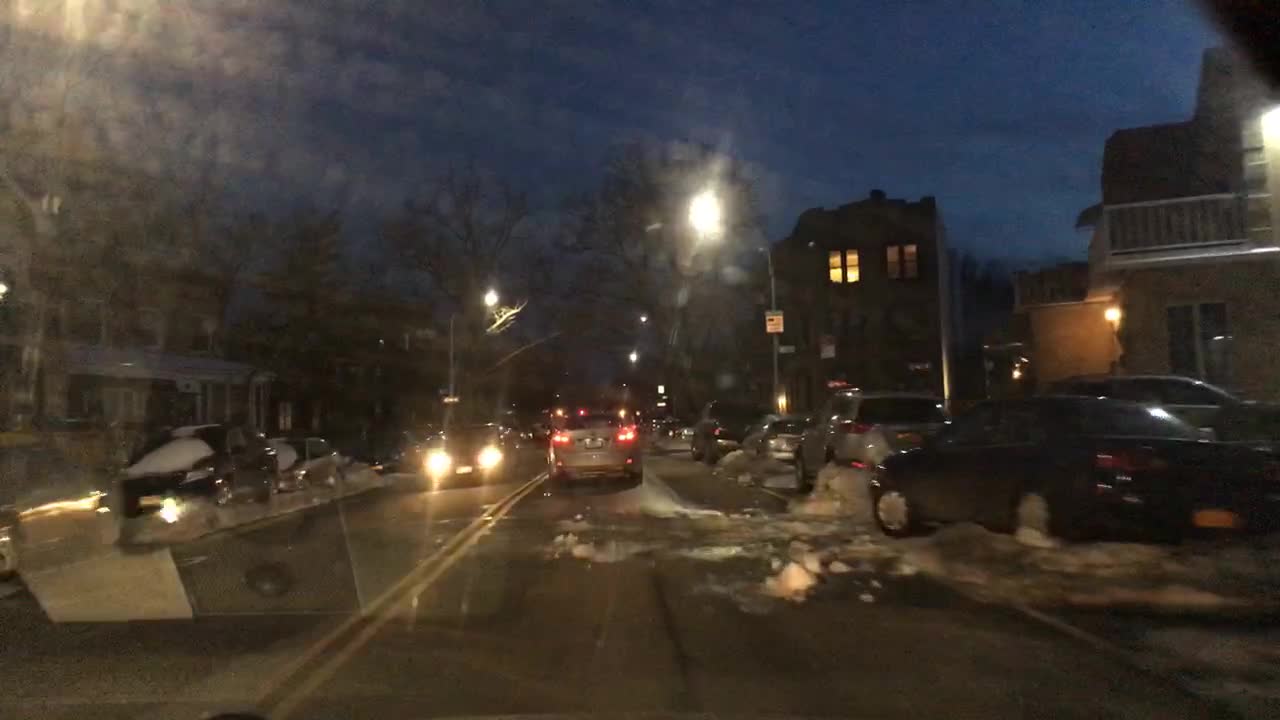}
\includegraphics[width=2cm]{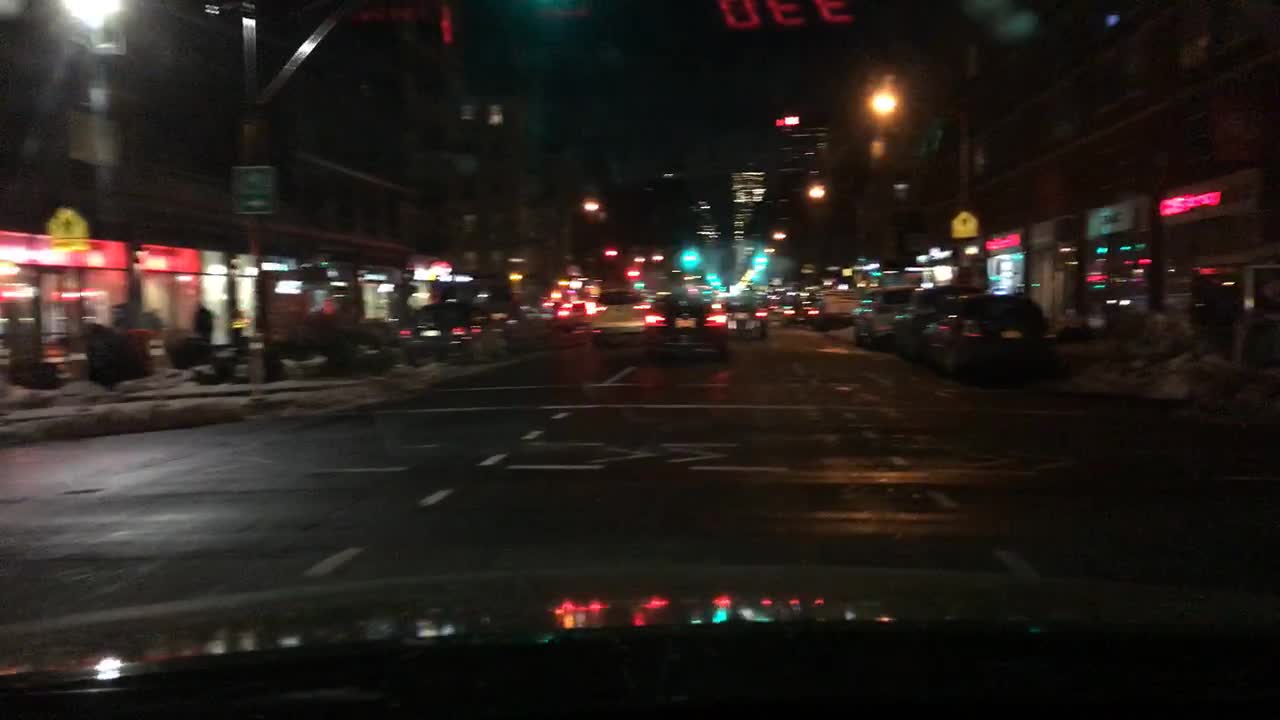}
\includegraphics[width=2cm]{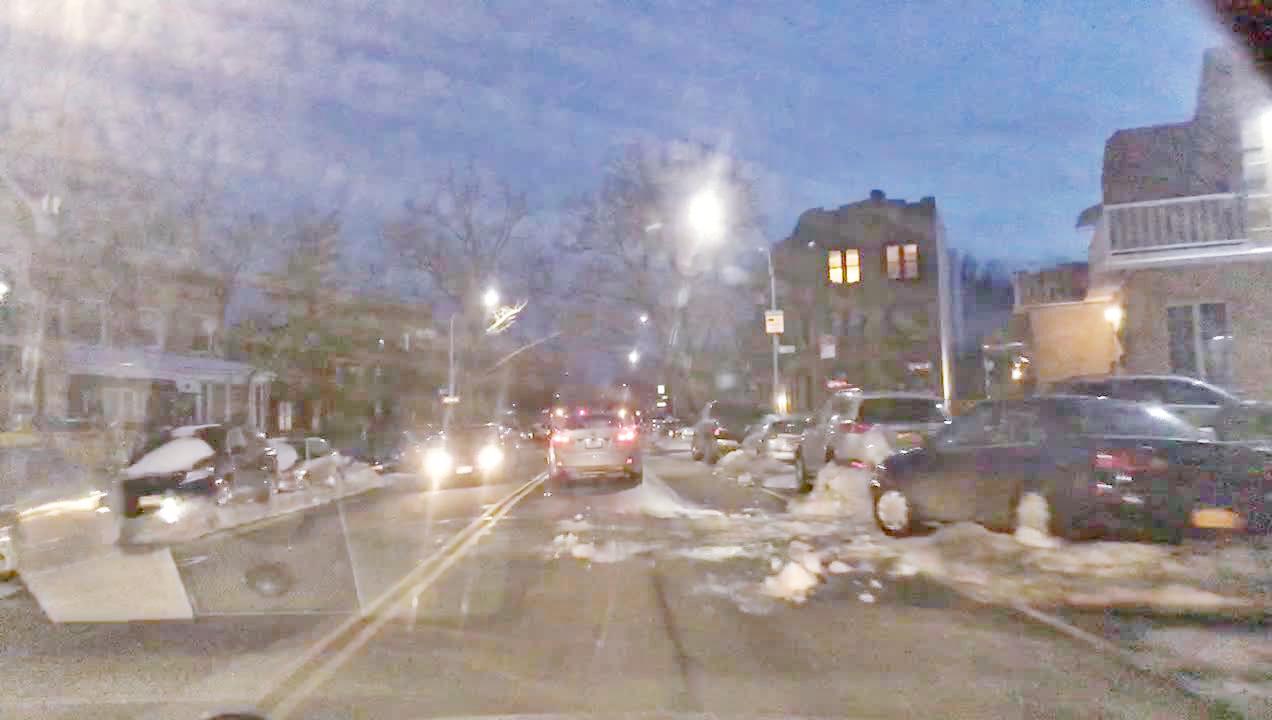}
\includegraphics[width=2cm]{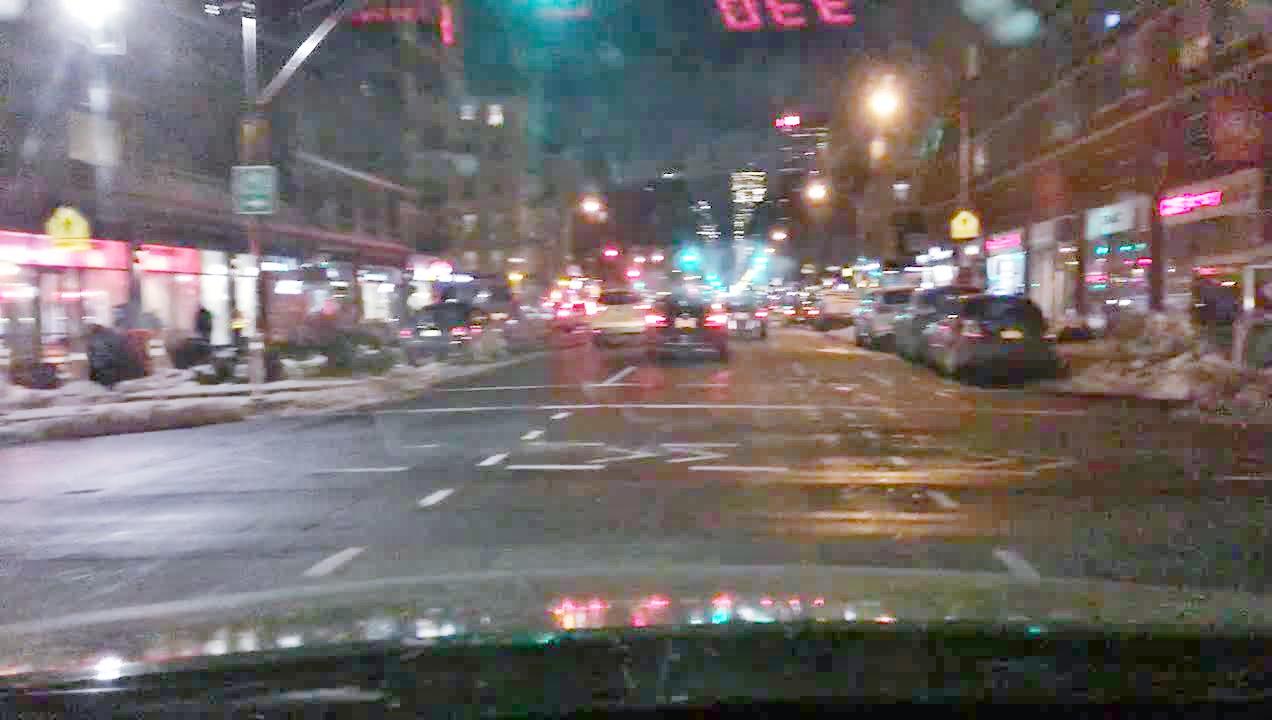}
\caption{The failure cases of the proposed model. Left two: low-light images. Right two: our enhanced results. Our model cannot address strong motion blurs or mirror reflection.}
\label{failure}
\end{figure}

\section{Conclusion}

This paper introduced a novel semantic-guided zero-shot low-light image enhancement network. The proposed network is trainable without paired images, unpaired datasets, or segmentation labels. That is achieved by enhancement factor extraction, recurrent image enhancement, and unsupervised semantic segmentation. Extensive experiments demonstrated the excellence of the proposed method in terms of perceptual quality, model efficiency, and the benefits for high-level vision tasks. Our future plan is to investigate motion blur removal with low-light image/video enhancement \cite{hu2014deblurring}. We also intend to explore detection-driven enhancement algorithms \cite{zheng2021deblur}.



\clearpage
\balance

{\small
\bibliographystyle{ieee_fullname}
\bibliography{egpaper}
}

\end{document}